\documentclass[10pt,twocolumn,letterpaper]{article}

\usepackage[pagenumbers]{cvpr} 
\usepackage{url}

\usepackage[utf8]{inputenc} 
\usepackage[T1]{fontenc}    
\usepackage{amsfonts}       
\usepackage{nicefrac}       
\usepackage{microtype}      
\usepackage{xcolor}         
\usepackage[linesnumbered,ruled,vlined]{algorithm2e}
\usepackage{array}
\usepackage{amssymb}
\usepackage{latexsym}
\usepackage{amsmath,amssymb}
\usepackage{graphicx}
\usepackage{multirow}
\usepackage{adjustbox}
\usepackage{makecell}
\usepackage[nottoc]{tocbibind}
\usepackage{graphicx}
\usepackage{kotex}
\usepackage{booktabs, multirow} 
\usepackage{nicematrix,caption}
\usepackage{soul}
\usepackage{changepage,threeparttable} 
\usepackage{subcaption}
\usepackage{color}

\newcommand{\add}[1] {\textcolor{blue}{#1}} 

\newcommand*{\V}[1]{\mathbf{#1}}
\newcommand*{\C}[1]{\mathcal{#1}}
\definecolor{cvprblue}{rgb}{0.21,0.49,0.74}
\usepackage[pagebackref,breaklinks,colorlinks,citecolor=cvprblue]{hyperref}

\def\paperID{xxxx} 
\def\confName{xxxx}
\def\confYear{xxxx}

\title{Detailed Human-Centric Text Description-Driven Large Scene Synthesis}

\author{Gwanghyun Kim$^{1*}$, Dong Un Kang$^{1*}$, Hoigi Seo$^{1*}$, Hayeon Kim$^{1*}$, \stepcounter{footnote} Se Young Chun$^{1,2}$\thanks{} \\
$^1$Dept. of Electrical and Computer Engineering, $^2$INMC \&  IPAI \\
Seoul National University, Republic of Korea \\
{\tt\small \{gwang.kim, qkrtnskfk23, seohoiki3215, khy5630,  sychun\}@snu.ac.kr}
}

\begin{document}


\begin{figure*}
\twocolumn[{
\maketitle
\begin{center}
\vspace{-2.em}
\includegraphics[width=0.97\textwidth]{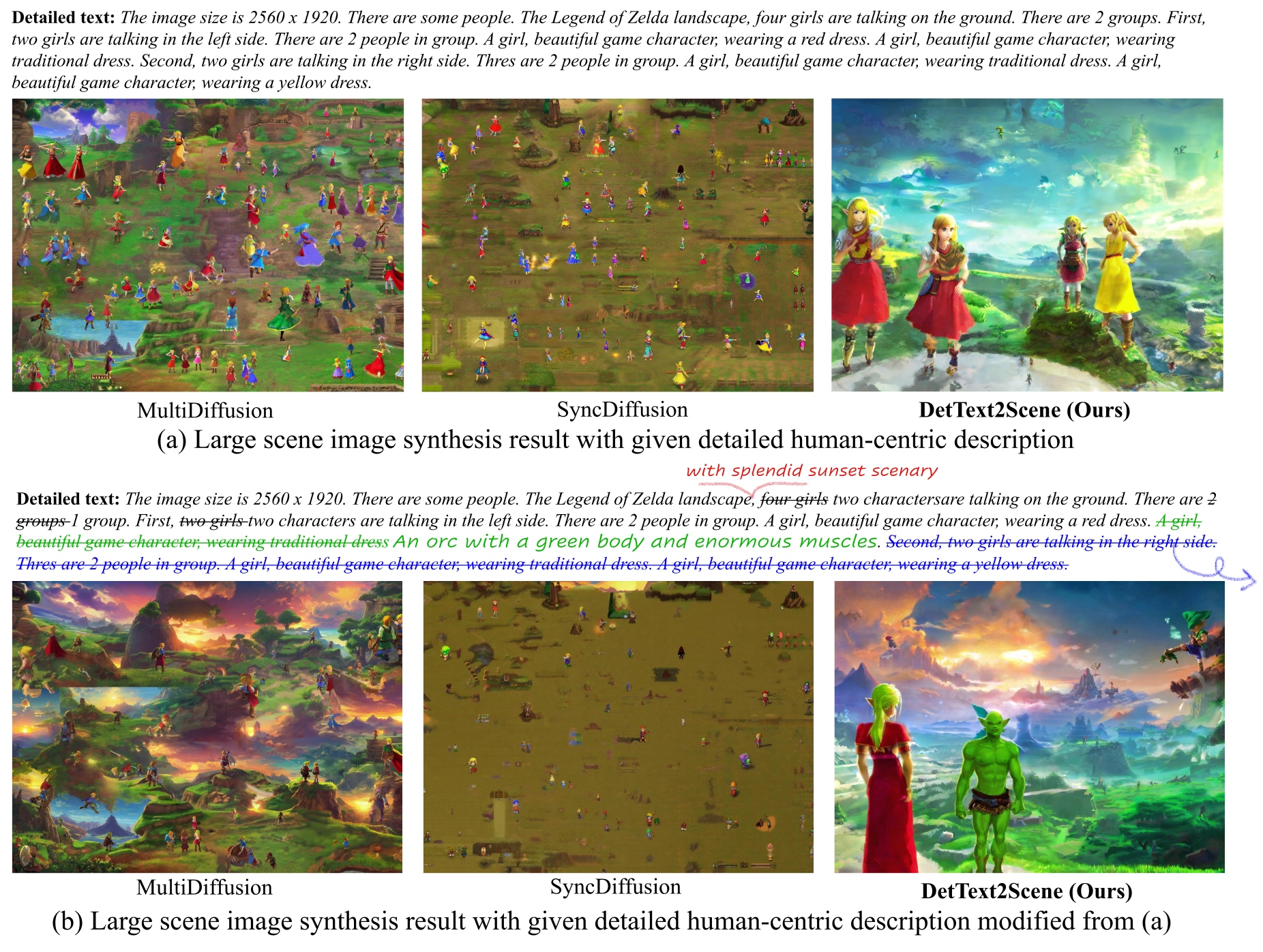}
     \vspace{-1.3em}
\caption{Large scene (2560$\times$1920) synthesis driven by (a) the detailed description and (b) its modified text with added (splendid sunset), changed (a girl to an orc), and removed (two girls on right) details. Compared to prior arts, MultiDiffusion~\citep{bar2023multidiffusion} and SyncDiffusion~\citep{lee2023syncdiffusion}, our DetText2Scene enables large scene generation with high \textit{faithfulness}, \textit{naturalness} and \textit{controllability} reflecting the given text descriptions. Note that all prior arts in this figure and our proposed method are based on Stable Diffusion 1.5~\citep{rombach2022high}.}
\label{fig1}
\vspace{-.5em}
\end{center}
}]
\end{figure*}

{
  \renewcommand{\thefootnote}%
    {\fnsymbol{footnote}}
    \footnotetext[1]{Authors contributed equally. \ $^\dagger$Corresponding author.}
  
}

\begin{abstract}{
Text-driven large scene image synthesis has made significant progress with diffusion models, but controlling it is challenging. While using additional spatial controls with corresponding texts has improved the controllability of large scene synthesis, it is still challenging to faithfully reflect detailed text descriptions without user-provided controls. Here, we propose DetText2Scene, a novel text-driven large-scale image synthesis with high faithfulness, controllability, and naturalness in a global context for the detailed human-centric text description. Our DetText2Scene consists of 1) hierarchical keypoint-box layout generation from the detailed description by leveraging large language model (LLM), 2) view-wise conditioned joint diffusion process to synthesize a large scene from the given detailed text with LLM-generated grounded keypoint-box layout and 3) pixel perturbation-based pyramidal interpolation to progressively refine the large scene for global coherence. Our DetText2Scene significantly outperforms prior arts in text-to-large scene synthesis qualitatively and quantitatively, demonstrating strong faithfulness with detailed descriptions, superior controllability, and excellent naturalness in a global context.
}\end{abstract}
    
\section{Introduction}
\label{sec:intro}

Recent advances in text-to-image generation with diffusion models trained with billions of images offer advantages not only in producing high-quality, realistic images~\citep{ramesh2022hierarchical,midjourney,rombach2022high}, but also in applying to various tasks such as image inpainting~\citep{lugmayr2022repaint, nichol2021glide, saharia2021image}, image editing~\citep{meng2021sdedit,kim2022diffusionclip,hertz2022prompt,brooks2022instructpix2pix,avrahami2023blended_latent,nichol2021glide}, and image deblurring~\citep{whang2022deblurring, chung2023dps}, 3D contents generation~\citep{lin2023magic3d,tang2023makeit3d,xu2023dream3d,kim2023podia,kim2023datid,seo2023ditto} with few-shot adaptation~\citep{kumari2023multiconcept, gal2023image,ruiz2022dreambooth}. In particular, some of the recent works in text-to-image synthesis have extended the versatility of the pre-trained text-to-image diffusion models to generate large-scale images without computationally expensive additional training on them~\citep{avrahami2023blended_latent,bar2023multidiffusion,lee2023syncdiffusion}.

Synthesizing a large scene from text is still challenging in text-to-image generation. Joint diffusion-based methods~\cite{bar2023multidiffusion, zhang2023diffcollage, lee2023syncdiffusion} were proposed for generating seamless montage of images by performing the reverse generative process across multiple views simultaneously while averaging the intermediate noisy images in the overlapped regions every step from a text prompt~\citep{bar2023multidiffusion,lee2023syncdiffusion}. However, considering such a large canvas, the controllability with the detailed description in text-driven large scene generation is a desirable option to have. Applying additional semantic segmentation maps with corresponding texts shows the possibility of controlling diffusion models to ground multiple objects and humans~\citep{bar2023multidiffusion}.

Unfortunately, the detailed human-centric text description without additional spatial controls does not seem to work well for the prior arts in text-to-large scene generation, struggling to generate large images that fully convey the detailed descriptions with the following critical issues, as illustrated in the foremost two columns (MultiDiffusion~\citep{bar2023multidiffusion}, SyncDiffusion~\citep{lee2023syncdiffusion}) of Figure~\ref{fig1}(a).
The prior arts seems to \textit{lack controllability with detailed texts} such as no control on the number of generated humans and objects. This is possibly because 
the joint diffusion process trained on small images (e.g., 512$\times$512) uses the same text prompt for all views, thus overly duplicating tiny objects and humans. 
Moreover, the publicly available text-to-image diffusion model seems to have \textit{low faithfulness with the detailed texts}, resulting in missing objects  and binding incorrect attributes~\cite{chefer2023attend}.
For small image synthesis, this issue has been addressed with attention map control~\cite{chefer2023attend}, grounding models~\cite{li2023gligen, kim2023dense} or additional keypoints~\citep{zhang2023adding}. For the case of large-scene synthesis, due to the joint diffusion process that overly duplicates instances, the chances of missing objects are relieved. However, the attribute binding issue still remains.  
Also the joint diffusion has weak dependencies across far apart views and thus yields disconnected scenes with \textit{poor naturalness in global context}.
Lastly, while using spatial controls like segmentation maps with grounded text description for large scene synthesis has been investigated, controlling it only with texts has been under-explored.

\begin{figure*}[!tbp]

    \vspace{-0.5em}
    \centering
    \includegraphics[width=0.94\textwidth]{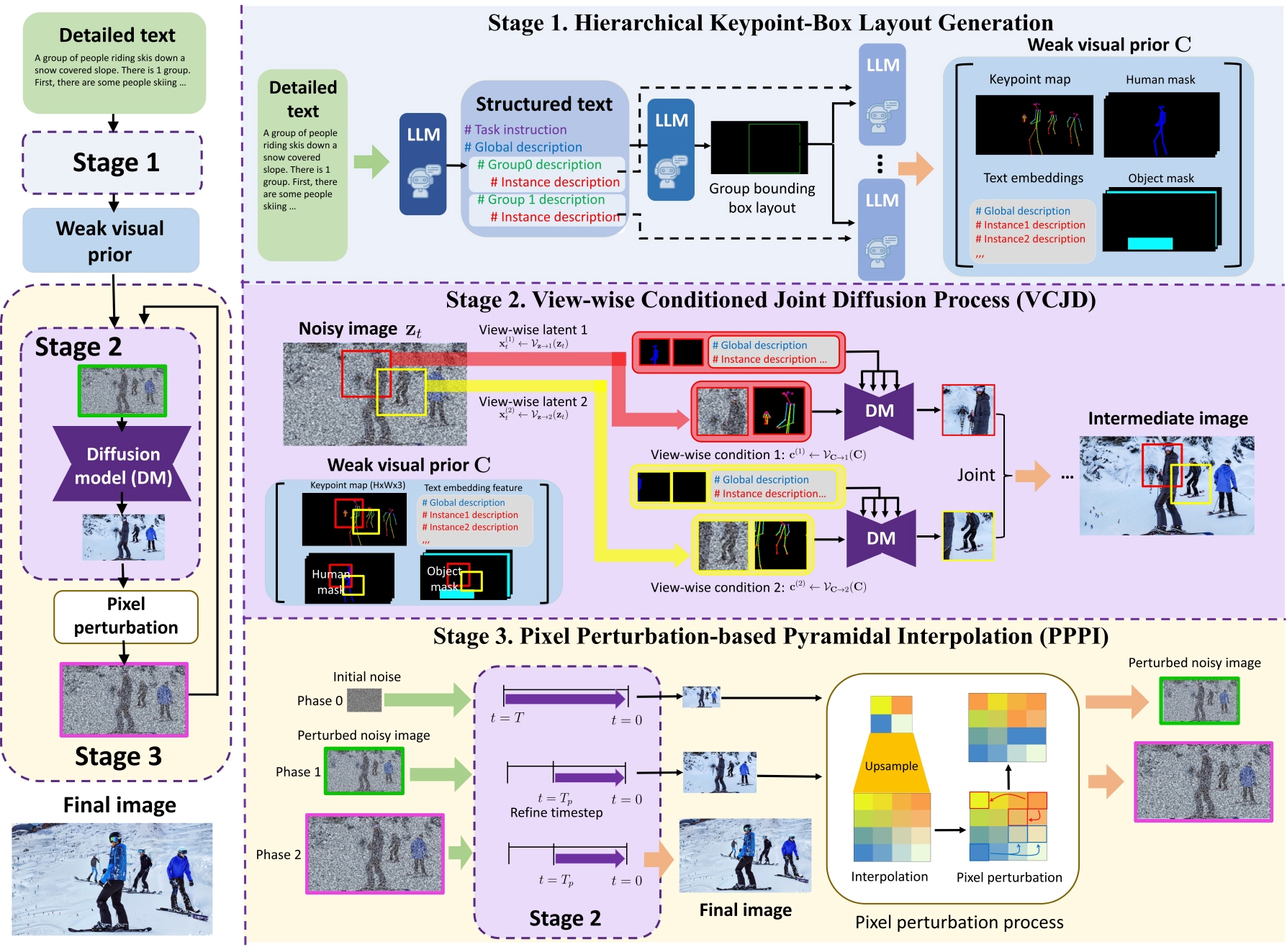}
    \vspace{-0.8em}

    \caption{Overall pipeline of our proposed DetText2Scene. \textbf{(Stage 1) Hierarchical keypoint-box layout generation:} From the detailed text description, the fine-tuned LLMs sequentially generate group boxes and keypoint-box layout of instances for weak visual priors. \textbf{(Stage 2) View-wise conditioned joint diffusion (VCJD):} With LLM-generated weak visual priors such as keypoint map, instance mask and text embedding feature, our VCJD jointly proceeds view-wise grounding for faithful and controllable image synthesis. \textbf{(Stage 3) Pixel perturbation-based pyramidal interpolation (PPPI):} To improve the global consistency of large scene generation, the PPPI recurrently interpolates the generated image with pixel perturbation process that incorporates VCJD.}
    \vspace{-1.3em}
    \label{fig23}
\end{figure*}

Here, we propose DetText2Scene (Detailed Text To Scene), a novel text-driven large scene synthesis method from detailed human-centric text description to generate \textit{controllable} and \textit{natural} large-scale images \textit{faithfully} corresponding to the texts without additionally provided spatial controls. Our DetText2Scene consists of 1) generating weak visual prior that embodies hierarchical keypoint and bounding box layout from detailed texts
by leveraging large language model (LLM) to outline several people and objects (Stage 1 in Figure~\ref{fig23}), 2) view-wise conditioned joint diffusion process (VCJD) for large scene synthesis, conditioned on spatial controls and ground attributes from generated weak visual priors
and detailed texts (Stage 2 in Figure~\ref{fig23}) and 3) pixel perturbation-based pyramidal interpolation (PPPI) (Stage 3 in Figure~\ref{fig23}) to recurrently refine the generated images for further improved quality and global consistency across views, overcoming weak dependencies across views.
{Our experiments demonstrated that our DetText2Scene enabled large-scene synthesis from detailed human-centric text with exquisite quality, achieving firm faithfulness with detailed descriptions (as assessed by CLIP score and >67.2\% of preference over existing methods in user study), strong controllability (as measured by human count accuracy), and high naturalness in both global context (>65.6\% of preference over existing methods in user study) and human instance quality (>65.3\% of preference over existing methods in user study). When keypoint-box layouts were given, our ablation studies demonstrate that our DetText2Scene still outperforms prior arts in text-to-scene generation.}

\section{Related Works}
\label{sec:RelatedWorks}

\subsection{Text-guided large scene synthesis via diffusion models}
Recent works~\cite{bar2023multidiffusion, zhang2023diffcollage, lee2023syncdiffusion} in text-guided large scene generation have conducted diffusion in multiple views jointly while combining noisy latent features or scores at each reverse diffusion step.
In specific, when we denote sampling the subsequent denoised data in the reverse process of diffusion models with the learned distribution $p_\theta (\V{x}_{t-1} | \V{x}_t)$ defined in the sampling methods~\cite{ho2020denoising, song2020denoising, liu2022pseudo} as $\V{x}_{t-1} = \C{D}(\V{x}_t, t, \epsilon)$ where $\epsilon \sim \mathcal{N}(0,\mathcal{I})$, we can consider the case of generating a arbitrary large-scene image $\V{z} \in \mathbb{R}^{H_z \times W_z \times D}$. 
The image at each window $\V{x}^{(i)} \in \mathbb{R}^{H_x \times W_x \times D}$ is a subarea of the large scene image whose union across all the windows covers the entire large-scene image.
Let ${m}^{(i)} \in [0, 1]^{H_x \times H_x}$ denote a binary mask for the subregion in the large-scene image corresponding to the $i$-th window. 
The function $\C{V}_{\V{z} \rightarrow i}: \mathbb{R}^{H_z \times W_z \times D} \rightarrow \mathbb{R}^{H_x \times W_x \times D}$ maps (crops) the large-scene image $\V{z}$ to the $i$-th window image, while $\C{V}_{i \rightarrow \V{z}}: \mathbb{R}^{H_x \times W_x \times D} \rightarrow \mathbb{R}^{H_z \times W_z \times D}$ is its inverse function that fills the region outside of the mask ${m}_i$ with zeros.
During the joint diffusion process running the reverse process simultaneously for each window, the noisy images from the windows ${\V{x}^{(i)}_t}$ are first averaged in the large scene space $\V{z}_{t} = \frac{\sum_i \C{V}_{i \rightarrow \V{z}} (\V{x}^{(i)}_t)}{\sum_i {m}^{(i)}}$.
These joint diffusion-based methods enable the generation of seamless, arbitrary-sized images. 
However, existing methods still struggle to generate large scene images that fully convey detailed descriptions. 
In this work, we propose DetText2Scene, which enables to synthesize large-scale images with significantly improved faithfulness, controllability, and naturalness.

\subsection{Text-guided image synthesis with spatial control via diffusion models}
Recently, several works~\citep{avrahami2023spatext,li2023gligen,qin2023unicontrol,mou2023t2i,zhang2023adding} propose to enable spatial control with weak visual prior
such as segmentation map~\citep{avrahami2023spatext}, depth map~\cite{li2023gligen,qin2023unicontrol,mou2023t2i, zhang2023adding}, dense caption~\cite{xie2023boxdiff, kim2023dense} and human keypoint~\cite{qin2023unicontrol,mou2023t2i, zhang2023adding} to a pre-trained text-to-image diffusion model.
For example, the reverse diffusion step for image synthesis guided by human keypoint pose map $k$  and text prompt $y$ will be $\V{x}_{t-1} = \C{D}(\V{x}_t, t, \epsilon, y, k)$.  
These works enable the achievement of more controllable synthesis.
However, existing text-to-large-scene methods face challenges in producing images with precise control.
In this work, we elongate the virtue of controllable synthesis to large scene synthesis with DetText2Scene only by text.

\subsection{Layout generation via large language models}
Several concurrent works~\citep{feng2023layoutgpt, xie2023visorgpt,cho2023visual} leverage the potential of large language models for enhancing text-to-image models. 
LayoutGPT~\citep{feng2023layoutgpt} leverages GPT to create layouts from text conditions and then generate images from created layouts.
VP-T2I~\cite{cho2023visual} finetunes an open-source language model for the specific text-to-layout task and uses standard layout-to-image models for image generation.
VisorGPT~\citep{xie2023visorgpt}  explicitly learns the probabilistic visual prior through generative pre-training. 
However, the methods of layout generation targeted for large-scene generation from detailed text description have been scarcely addressed. 
To achieve this, we generate a keypoint-box layout simultaneously and create a hierarchical pipeline to generate complex scenes.

\section{Method: DetText2Scene}
\label{sec:DetText2Scene}

We first generate a keypoint-box layout of the scene from the detailed human-centric description, which provides spatial controls for the image generation process, as represented in the stage 1 of Fig~\ref{fig23}. 
Then, we propose a view-wise conditioned joint diffusion process to synthesize a large scene from the given detailed text and the LLM-generated spatial controls in grounded keypoint-box layouts as depicted in the stage 2 of Fig~\ref{fig23}.
Lastly, we propose the pixel perturbation-based pyramidal interpolation to ensure the high quality and global dependencies across the images as illustrated in the stage 3 of Fig~\ref{fig23}.

\subsection{Hierarchical keypoint-box layout generation}
\label{3.1}
In large-scene generation from detailed human-centric description, it is critical to maintain a global context while creating multiple instance layouts. To yield this, we proposed hierarchical pipeline that sequentially deals with weak visual prior from groups to instances as coarse-to-fine framework as represented in the stage 1 of Fig.~\ref{fig23}.

\paragraph{Hierarchical keypoint-box layout dataset.}
We generated the hierarchical keypoint-box layout labeled with detailed descriptions based on the CrowdCaption dataset~\cite{wang2022happens}.
As shown in Fig.~\ref{fig23}, the detailed input text is converted to hierarchical structured prompts, including the quantity and descriptions of each global, group and instance. To construct the hierarchical keypoint-box layout dataset for instruction tuning, we firstly extracted dense captions and visual location of each instances using existing models of dense captioning (GRiT~\citep{wu2022grit},ViT-B), image captioning (BLIP~\citep{li2022blip},ViT-B) and pose estimation (ViTPose-B~\citep{xu2022vitpose}) which is based on the bounding box of instances. Then, for aligning process of acquired information, the dense caption and keypoint of instances are paired by intersection over union (IoU) matching. We checked the center of instance boxes to which group box it belongs and matched them with group grounding annotation. The details of the hierarchical keypoint-box layout dataset are described in the supplementary.

\paragraph{Instruction tuning for hierarchical keypoint-box layout generation.}
We propose hierarchical keypoint-box layout generation by fine-tuning the pre-trained LLM~\citep{touvron2023llama} to construct the weak visual priors structured from global outline to local keypoint-box layout of the image. First, we fine-tuned the LLM to convert the detailed natural texts to hierarchical structured text, called Nat2Hier LLM. Second, we fine-tuned the global grounding LLM to generate the proper group box layout from the global and group description prompt. 
Then, the local grounding LLM is fine-tuned to generate the keypoint-box layout of each instance from hierarchical descriptions and generated group box layout prompt. The examples of instruction-answer pairs of each grounding LLM are illustrated in the supplementary.

\subsection{View-wise conditioned joint diffusion}
\label{sec:process}
We enable to synthesize a large scene from the given detailed text and the spatial controls in grounded keypoint-box layout through view-wise conditioned joint diffusion process (VCJD) as illustrated in the stage 2 of Fig.~\ref{fig23} and Algorithm~\ref{algo_vcjd}.

\setlength{\floatsep}{0.em}
\setlength{\textfloatsep}{0.2em}

\begin{algorithm}[!b]
\caption{Large scene generation with view-wise conditioned joint diffusion process (VCJD)}
\label{algo_vcjd} 
{
    \footnotesize

    \SetKwFunction{FVCJDFEV}{VCJD-step}
    \SetKwProg{Fn}{Function}{:}{}
    \Fn{\FVCJDFEV{$\{\V{x}^{(i)}_{t}\}, t, \{\V{c}^{(i)} \}$}}{
        \add{\scriptsize \tcp{Denoise each view-wise latent}}
        \For{$i = 0, \dots, M-1$} {
             $\V{\tilde{x}}^{(i)}_{t-1} \leftarrow  \C{D}(\V{x}_t^{(i)}, t, \epsilon,  \V{c}^{(i)})$  \; 
        }
        \add{\scriptsize \tcp{Stitch latent by averaging globally}}
        $\V{z}_{t-1} \leftarrow \frac{\sum_i \C{V}_{i \rightarrow \V{z}} (\V{\tilde{x}}^{(i)}_{t})}{\sum_i {m}^{(i)}} $\;
        \add{\scriptsize \tcp{View-wise latent generation}}
        \For{$i = 0, \dots, M-1$} {
            $\V{x}^{(i)}_{t-1} \leftarrow \C{V}_{\V{z} \rightarrow i}(\V{z}_{t-1})$\;
        }
        \KwRet $\{ \V{x}^{(i)}_{t-1} \}$
    }
    \BlankLine

    \SetKwFunction{FVCJD}{VCJD}
    \Fn{\FVCJD{$\V{z}_{T_b}, \V{C}, T_b$}}{
        \add{\scriptsize \tcp{View-wise latent \& condition generation}}
        $\{ \V{x}^{(i)}_{T_b} \}_{i=0 \cdots M-1} \leftarrow  \{\C{V}_{\V{z} \rightarrow i}(\V{z}_{T_b})\}_{i=0 \cdots M-1} $\;
        $\{ \V{c}^{(i)} \}_{i=0 \cdots M-1} \leftarrow \{\C{V}_{\V{C} \rightarrow i}(\V{C})\}_{i=0 \cdots M-1} $\;

        \add{\scriptsize \tcp{Run VCJD process}}
        \For{$t = T_b, \dots, 1$} {
                $\{ \V{x}^{(i)}_{t-1} \}  \leftarrow$  \FVCJDFEV{$\{\V{x}^{(i)}_{t}\}, t, \{\V{c}^{(i)} \}$}\;
            }
    
        $\V{z}_{0} \leftarrow \frac{\sum_i \C{V}_{i \rightarrow \V{z}} (\V{\tilde{x}}^{(i)}_{0})}{\sum_i {m}^{(i)}}$ \;

        \KwRet $\V{z}_{0}$
    } 
}

\end{algorithm}

\paragraph{View-wise condition generation.}
From the stage 1, we could obtain the global condition set $\V{C}$, which is composed of the structured text prompts $\V{Y}$, the global human-object bounding boxes $\V{B}$, and the global human keypoint map $\V{K}$.
Here, we introduce the function $\C{V}_{\V{C} \rightarrow i}$ that maps the global condition $\V{C}$ into the $i$-th view-wise condition $\V{c}^{(i)}$, as the function $\C{V}_{\V{z} \rightarrow i}$ maps the large-scene image $\V{z}$ to the $i$-th patch $\V{x}^{(i)}$ as used in the joint diffusion process~\cite{bar2023multidiffusion, lee2023syncdiffusion}. 
In specific, $\V{c}^{(i)}$ consists of the view-wise full text $y^{(i)}$, human keypoint $k^{(i)}$ and dense caption pairs $\{(y^{(i)}_n,s^{(i)}_n)\}_{n=0 \cdots N-1}$ where $y_n^{(i)}$ is a non-overlapping description of each human or object that is a part of $y$, and $s_n^{(i)}$ denotes a binary map extracted from box or pose mask layout.

\paragraph{Dense keypoint-box text-to-image diffusion.}
Although several methods have explored keypoint-driven diffusion synthesis~\cite{qin2023unicontrol, mou2023t2i, zhang2023adding}, they exhibit limitations in  conditioning a single text on all human instances in the image, leading to a lack of controllability.
To achieve meticulous control over human instances, including factors like gender, clothing color, and character attributes, we introduce a novel approach called dense keypoint-box text-to-image diffusion. 
This method involves extending current keypoint-based diffusion models~\cite{zhang2023adding} to ground diverse attributes and generate grounded objects. This is achieved through the incorporation of attention modulation~\citep{kim2023dense}, a technique that dynamically adjusts attention scores between image segments and text segments.

The reverse diffusion step from the noisy latent $\V{x}_t$ at the diffusion timestep $t$ and the condition set $ \V{c}$ can be represented as:
\small
\begin{align}
\V{x}_{t-1} =  \C{D}(\V{x}_t, t, \epsilon, \V{c}) =  \C{D}(\V{x}_t, t, \epsilon,y, k,\{(y_n,s_n)\}_{n=0}^{N-1}),
\label{eq:dense}
\end{align}
\normalsize
where $\V{c}$ is composed of the full text $y$, keypoint $k$ and dense caption pairs $\{(y_n,s_n)\}_{n=0 \cdots N-1}$ and $\epsilon \sim \mathcal{N}(0,\mathcal{I})$.
See the supplementary for the further details of keypoint-based diffusion models~\cite{zhang2023adding} and the attention modulation methods~\citep{kim2023dense}.

\paragraph{Large scene image generation with VCJD.}
Building upon these components, VCJD facilitates the controlled synthesis of a large scene based on detailed human-centered text and spatial controls.
This process avoids introducing artifacts such as excessive duplication of objects and humans, while also preventing the generation of overly small elements.
VCJD initiates the synthesis process by starting with random noise or a noisy latent vector $\V{z}_{T_b}$ at timestep $T_b$ to begin with, alongside the global condition $\V{C}$. 
The image is then systematically generated through a view-wise conditioned joint reverse diffusion process, outlined in Algorithm~\ref{algo_vcjd}.

\subsection{Pixel perturbation-pyramidal interpolation}
\label{sec:process}
In existing joint diffusion-based methods, maintaining global coherence across distant views poses a challenge, leading to unrealistic results. 
To adress this, we propose pixel perturbation-based pyramidal interpolation (PPPI) as described in the stage 3 of Fig.~\ref{fig23} and Algorithm~\ref{algo_pppi}.
Our method achieves the generation of highly detailed and naturally coherent large scenes through novel pixel perturbation, requiring minimal computational resources. This stands in contrast to cascaded Diffusion~\cite{ho2021cascaded}, which relies on super-resolution models, introducing additional complexity and computational demands.

\begin{algorithm}[!t]
\caption{Pixel perturbation-based pyramidal interpolation (PPPI)}
\label{algo_pppi} 
{
    \footnotesize
    \KwIn{$\V{z}_{T}^0 \sim \mathcal{N}(0,\mathcal{I}), \V{C}, \alpha_{\textnormal{interp}}, \gamma_{\textnormal{pert}}, , d_{\textnormal{pert}}$}
    \KwOut{$\V{z}_{0}^P$}

    \SetKwFunction{FPPPI}{PPPI}
    \SetKwFunction{FPP}{PixelPert}
    \SetKwFunction{FINTERP}{Interpolation}
    \SetKwFunction{FADD}{Forward}
    \SetKwFunction{FRAND}{Rand}
    \SetKwProg{Fn}{Function}{:}{}
    
    \Fn{\FPP{$\V{z}_{0}^{p}$, $\V{\bar{z}}_{0}^{p+1}$, $\alpha_{\textnormal{interp}}$,  $ \gamma_{\textnormal{pert}}$, $d_{\textnormal{pert}}$}}{
        \For{$h = 0, \dots, H-1$} {
            \For{$w = 0, \dots, W-1$} {
                \If{$(prob \sim \C{U}[0, 1])\textgreater p_{pert}$}{
                    $ H_{rand} = h/\alpha + $\FRAND{$- d_{\textnormal{pert}},  d_{\textnormal{pert}}$}\;
                    $ W_{rand} = w/\alpha + $\FRAND{$- d_{\textnormal{pert}}, d_{\textnormal{pert}}$}\;
                    $\V{\bar{z}}_{0}^{p+1}[h, w] = \V{z}_{0}^{p}[H_{rand}, W_{rand}]$\;
                    }
                }
            }
        \KwRet $\V{\bar{z}}_{0}^{p+1}$
        }
    
    \BlankLine
    \footnotesize
    \add{\scriptsize \tcp{Initial image generation}}
    $\V{z}_{0}^0  \leftarrow$ \FVCJD{$\V{z}^{0}_{T_0}, \V{C}, T_0$}\;
    \add{\scriptsize \tcp{Pyramidal image generation}}
    \For{$p$ = $0, \dots, P-1$} {
        $\V{\bar{z}}_{0}^{p+1} \leftarrow$ \FINTERP{$\V{z}_{0}^{p}, \alpha_{\textnormal{interp}}$}\;
        $\V{\tilde{z}}_{0}^{p+1} \leftarrow$ \FPP{$\V{z}_{0}^{p}, \V{\bar{z}}_{0}^{p+1},\alpha_{\textnormal{interp}}, \gamma_{\textnormal{pert}}, d_{\textnormal{pert}}$}\;
        $\V{z}_{T_{p+1}}^{p+1} \leftarrow$ \FADD{$\V{\tilde{z}}_{0}^{p+1},  T_{p+1}$}\;
        $\V{z}_{0}^{p+1}  \leftarrow$ \FVCJD{$\V{z}^{p+1}_{T_{p+1}}, \V{C}, T_{p+1}$}\;
    }
}
\end{algorithm}

\begin{figure}[!t]
\begin{center}
    \includegraphics[width=0.5\textwidth]{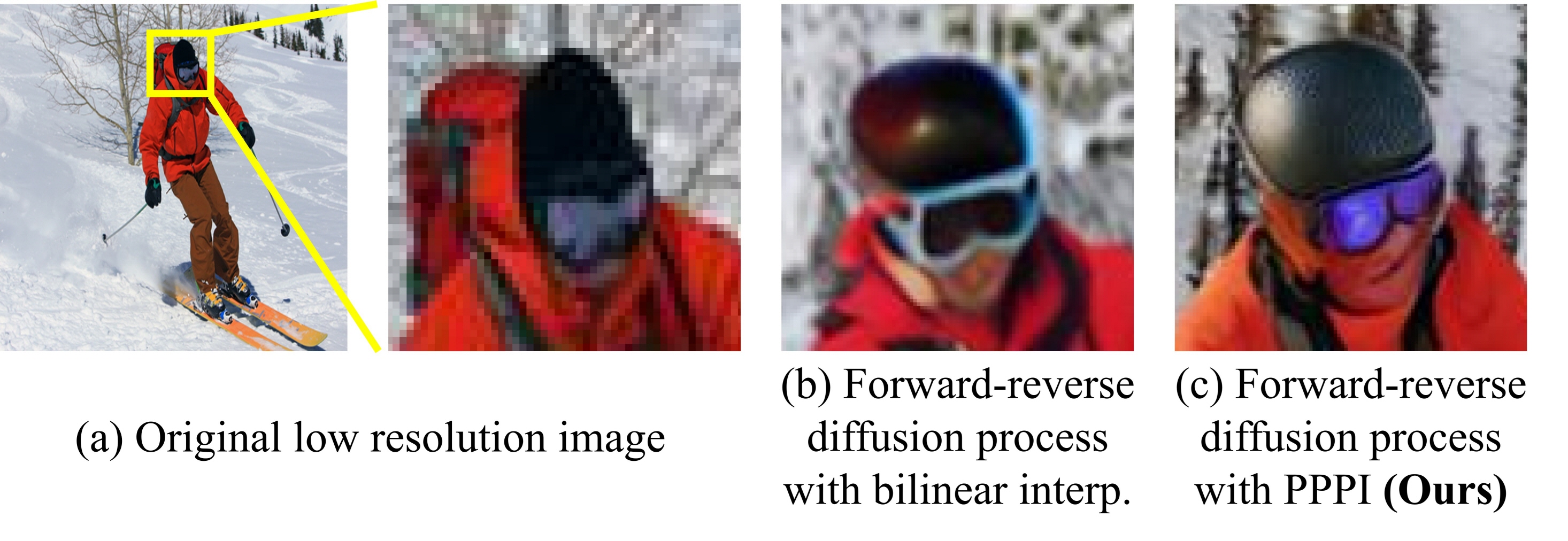}
\vspace{-2.5em}
    \caption{Examples of generated image after forward-reverse process using different interpolation methods with the description \textit{``a person riding skis"}. Our PPPI enable to efficiently generate sharper images and introduces more significant changes corresponding to the given description.
    }
    \label{fig3}
   \vspace{-.5em}
\end{center}
\end{figure}

\paragraph{Pyramidal image generation.}
Our pyramidal image generation method progressively synthesizes images from small to large scales to achieve a globally harmonious result. Starting with random initial noise $\V{z}_T^0$ and a global condition $\V{C}$, we generate a small but highly consistent image $\V{z}_0^p = \V{z}_0^0$ at the first phase $p=0$.
Subsequently, we upscale the image to $\V{\bar{z}}^{p+1}_{0}$ using a bilinear interpolation with a scaling factor $\alpha_{\textnormal{interp}}$.
Next, we apply a forward process for a specified refinement time step $T_{p+1}$, resulting $\V{z}^{p+1}_{T_{p+1}}$.
This is followed by a VCJD reverse process, generating $\V{z}^{p+1}_0$.
These image interpolation and forward-reverse processes are repeated until the desired image $\V{z}_0^P$ is obtained at the final phase $P$.

\begin{figure*}[!h]
    \centering
    \vspace{-1.5em}
    \includegraphics[width=0.95\textwidth]{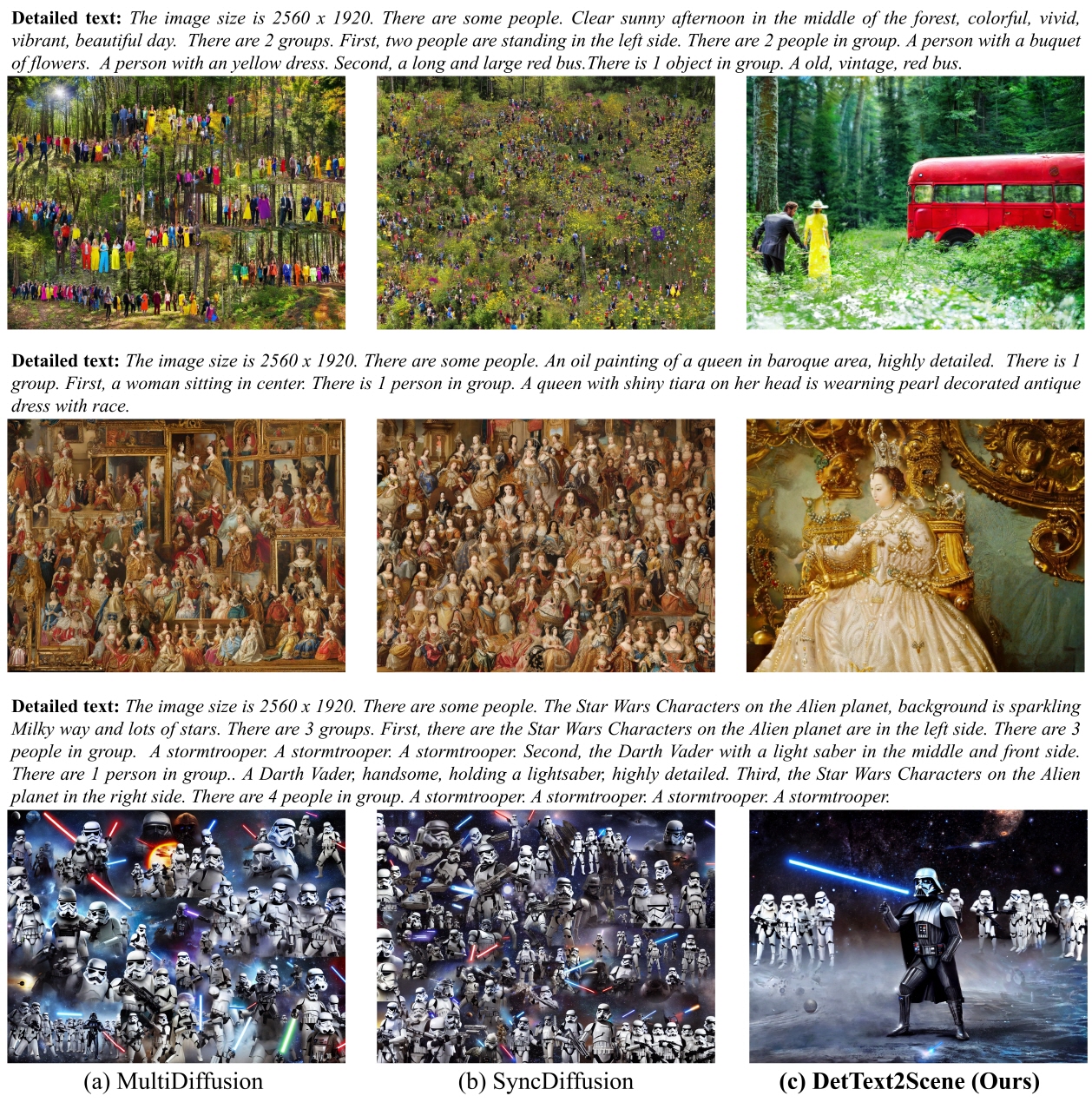}
    \vspace{-1em}
\caption{Qualitative results of large-scene image generation from the detailed text only. Compared to prior works, MultiDiffusion~\citep{bar2023multidiffusion} in the column (a) and SyncDiffusion~\citep{lee2023syncdiffusion} in the column (b), the results of our DetText2Scene in the column (c) show high \textit{faithfulness}, \textit{naturalness}, and \textit{controllability}. All the images were generated with the size of (2560$\times$1920). The Stable Diffusion 1.5~\citep{rombach2022high} is used for all the methods.}
     \label{fig4}
 \end{figure*}

\begin{table*}[!tp]
           \centering
           \captionsetup[subtable]{position = below}
          \captionsetup[table]{position=top}
           \vspace{-.5em}
           \begin{subtable}{0.4\linewidth}
               \centering
                \begin{adjustbox}{width=\linewidth}
\begin{tabular}{ccccc}\toprule
\multirow{2}{*}{\textbf{\makecell{Quantitative \\ result}}} &\multirow{2}{*}{CLIP score↑} &\multicolumn{3}{c}{$N_{\text{human}}$ matching} \\\cmidrule{3-5}
& &Prec.↑ &Rec↑ &F1↑ \\\midrule
MultiDiffusion &17.400 &0.233 &0.993 &0.378 \\
SyncDiffusion &32.017 &0.269 &0.982 &0.422 \\
\textbf{Ours} &32.097 &0.848 &0.933 &0.889 \\
\bottomrule
\end{tabular}
                \end{adjustbox}
           \end{subtable}%
           \hspace*{2em}
           \begin{subtable}{0.38\linewidth}
               \centering
                \begin{adjustbox}{width=\linewidth}
\begin{tabular}{ccccc}\toprule
\multirow{3}{*}{\textbf{User study}} &\multicolumn{3}{c}{Preference of ours ↑ } \\\cmidrule{2-4}
&\multirow{2}{*}{Faithfulness} &\multicolumn{2}{c}{Naturalness} \\
& &Global &Human \\\midrule
vs MultiDiffusion &67.2\% &65.6\% &65.3\% \\
vs SyncDiffusion &71.6\% &70.2\% &70.2\% \\
\bottomrule
\end{tabular}

                \end{adjustbox}
           \end{subtable}
\vspace{-0.5em}
           \caption{{Quantitative evaluation of large image generation from the detailed text. Our proposed DetText2Scene outperformed prior arts in multiple metrics and user studies.}}\label{tab1}
       \end{table*}

\paragraph{Pixel perturbation-based interpolation.}
In pyramidal image generation, processing the forward-reverse diffusion with a naively interpolated image leads to blurry output. 
This occurs because the diffusion models' learned distribution assigns high likelihood to the blurred image, resulting in minimal image modification.
We propose a pixel perturbation technique that forcefully injects high-frequency components into the interpolated image. This involves replacing adjacent pixels with a probability of $\gamma_{\textnormal{pert}}$
and a distance factor of $d_{\textnormal{pert}}$.
The proposed perturbation technique is illustrated in Algorithm~\ref{algo_pppi} and its effectiveness is demonstrated in Fig.~\ref{fig3}.

\begin{figure*}[!t]
     \centering
    \includegraphics[width=0.9\textwidth]{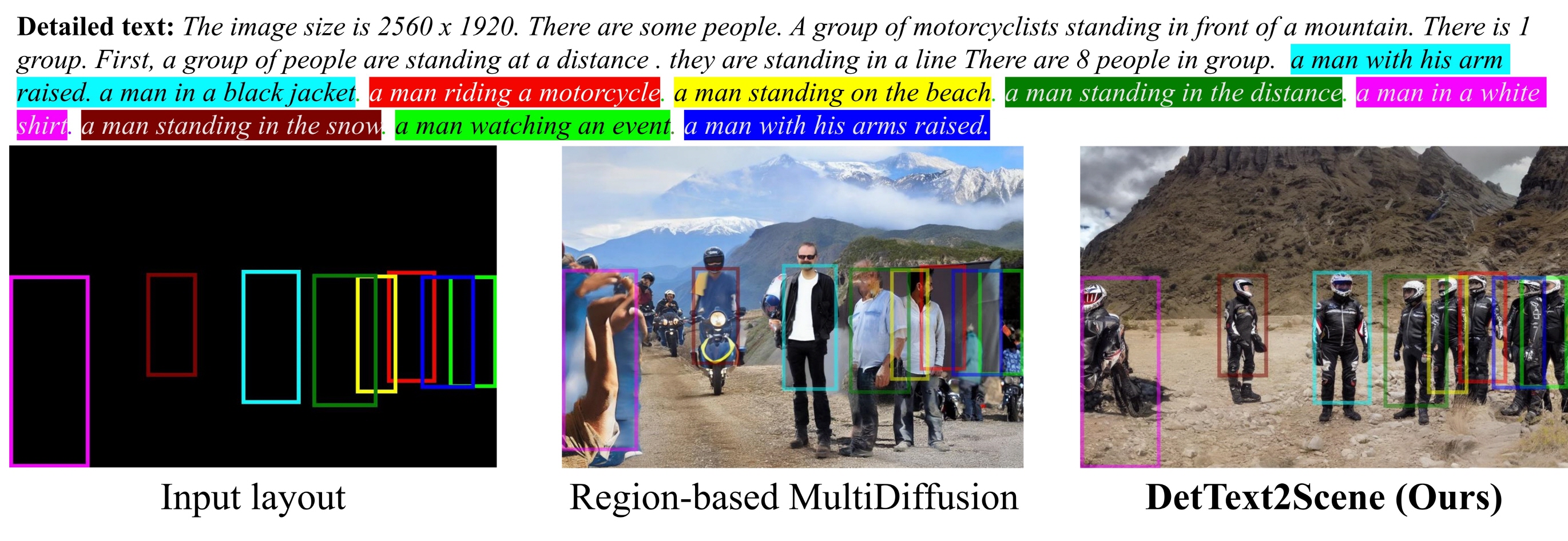}
    \vspace{-1em}
    \caption{Qualitative results of large-scene image generation from the spatial controls with detailed text. In the Region-based MultiDiffusion~\cite{bar2023multidiffusion}, bounding box masks with instance descriptions were employed to ground each instance. For better visualization, we matched the bounding box with corresponding instance description using distinct colors. The images were generated with the size of (1280$\times$960).}
    \label{fig5}
\end{figure*}

\begin{table*}[!tp]
           \centering
           \captionsetup[subtable]{position = below}
          \captionsetup[table]{position=top}
           \begin{subtable}{0.4\linewidth}
               \centering
                \begin{adjustbox}{width=\linewidth}
\begin{tabular}{cccccc}\toprule
\multirow{2}{*}{\textbf{\makecell{Quantitative \\ result}}} &\multirow{2}{*}{CLIP score↑} &\multicolumn{3}{c}{$N_{\text{human}}$ matching} \\\cmidrule{3-5}
& &Prec.↑ &Rec↑ &F1↑ \\\midrule
R-MultiDiffusion &20.045 &0.430 &0.986 &0.599 \\
\textbf{Ours} &27.220 &0.847 &0.932 &0.888 \\
\bottomrule
\end{tabular}
                \end{adjustbox}
           \end{subtable}%
           \hspace*{2em}
           \begin{subtable}{0.44\linewidth}
               \centering
                \begin{adjustbox}{width=\linewidth}
\begin{tabular}{ccccc}\toprule
\multirow{3}{*}{\textbf{User study}} &\multicolumn{4}{c}{Preference of ours ↑ } \\\cmidrule{2-5}
&\multirow{2}{*}{Faithfulness}  &\multirow{2}{*}{\shortstack[c]{Spatial \\ coherence}}&\multicolumn{2}{c}{Naturalness} \\
& & &Global &Human \\\midrule
vs R-MultiDiffusion &64.9\% & 62.4\% &71.5\% &69.4\% \\
\bottomrule
\end{tabular}
                \end{adjustbox}
           \end{subtable}
\vspace{-0.8em}
 \caption{Quantitative evaluation of image generation part for the given input layout and corresponding grounded text. 
 Our proposed DetText2Scene outperformed R-MultiDiffusion~\citep{bar2023multidiffusion} in multiple metrics and user studies.}\label{tab3}
 \vspace{-1em}
       \end{table*}

\section{Experiment}
\label{sec:Experiment}

 All previous works in the evaluation and our proposed method are based on Stable Diffusion 1.5~\citep{rombach2022high}. For dense keypoint-box text-to-image diffusion, we deploy ControlNet1.1~\citep{zhang2023adding} on the Stable Diffusion 1.5.
 For instruction tuning, we utilized the open-source toolkit of training LLMs~\citep{gao2023llamaadapterv2,zhang2023llamaadapter} for single-turn and multi-turn models. For efficient usage of resources, the quantized parameter efficient fine-tuning is adopted. 
Pixel perturbation swaps each pixel in the interpolated image with the existing pixel in $d_{\text{pert}}=1$ with $\gamma_{\text{pert}}=0.05$. 
See the supplementary for further details on our implementation and experiments and more results such as the large size of images and the evaluation of keypoint-box layout generation.

\subsection{Large scene synthesis from detailed text}
\label{subsec_eval1}
We begin by evaluating our method for large scene generation from detailed text that is a combination of our layout generation part and image generation part, which we targeted as a primary task.
We thoroughly evaluated our method with the state-of-the-art text-guided large scene generation methods as baselines, MultiDiffusion~\citep{bar2023multidiffusion} and SyncDiffuison~\citep{lee2023syncdiffusion} which show promising results by using joint diffusion process with the versatility of pre-trained diffusion models.
For quantitative comparisons and user study, we randomly select 100 real-world complex scene images in the test set of the hierarchical keypoint-box layout dataset, as described in the Section~\ref{3.1}.
For qualitative comparisons and user study, we create new detailed descriptions, making the above description more complex, applying various artistic styles and appearances on humans and objects to demonstrate the versatility of our method.
\paragraph{Qualitative comparisons.}
Our method exhibits exceptional quality, as evident in Fig.~\ref{fig4}.
For instance, in the first row of Fig.~\ref{fig4}, the baselines miss a large bus, whereas ours faithfully captures it. Our results also demonstrate superior binding, as exemplified by the distinct yellow and red colors for the dress and bus, as well as the accurate depiction of two individuals. In contrast, the baselines exhibit excessive duplication of humans and objects, suggesting a lack of controllability. Moreover, the baselines' results appear awkward and disconnected in the global context, and the human instance exhibits an unnatural physical structure. Conversely, our results maintain a natural appearance across the entire image.

\paragraph{Quantitative comparisons.}
We employ the global CLIP score~\citep{hessel2021clipscore} to evaluate the faithfulness of the generated images and the accuracy of the number of human instances($N_{\text{human}}$ matching) between the text prompt and the generated images. This metric, inspired by LayoutGPT's evaluation metric~\cite{feng2023layoutgpt}, assesses the controllability of the methods.
Our method surpasses the baselines, achieving a higher CLIP score and F1 scores for $N_{\text{human}}$ matching, as shown in Table~\ref{tab1}. The high recall values but critically low precision values  of MultiDiffusion~\citep{bar2023multidiffusion} and SyncDiffuison~\citep{lee2023syncdiffusion} indicate excessive duplication of objects and humans.

\paragraph{User study.}
As part of our research, we conducted a user study to assess the faithfulness and naturalness of images generated by MultiDiffusion~\citep{bar2023multidiffusion}, SyncDiffusion~\citep{lee2023syncdiffusion} and our DetText2Scene. Large-scene images were presented to participants, then asked to rank the methods based on the following criteria: 1) faithfulness to the text without missing objects or incorrect binding between words and objects, 2) naturalness from a global context, and 3) naturalness from a physical structure perspective.
As shown in Table~\ref{tab1}, out of the 9,045 responses from 201 participants,  our method was significantly preferred, with 65.6\%-71.6\% of the votes in all three criteria.

\begin{figure}[!t]
    \centering
    \includegraphics[width=0.45\textwidth]{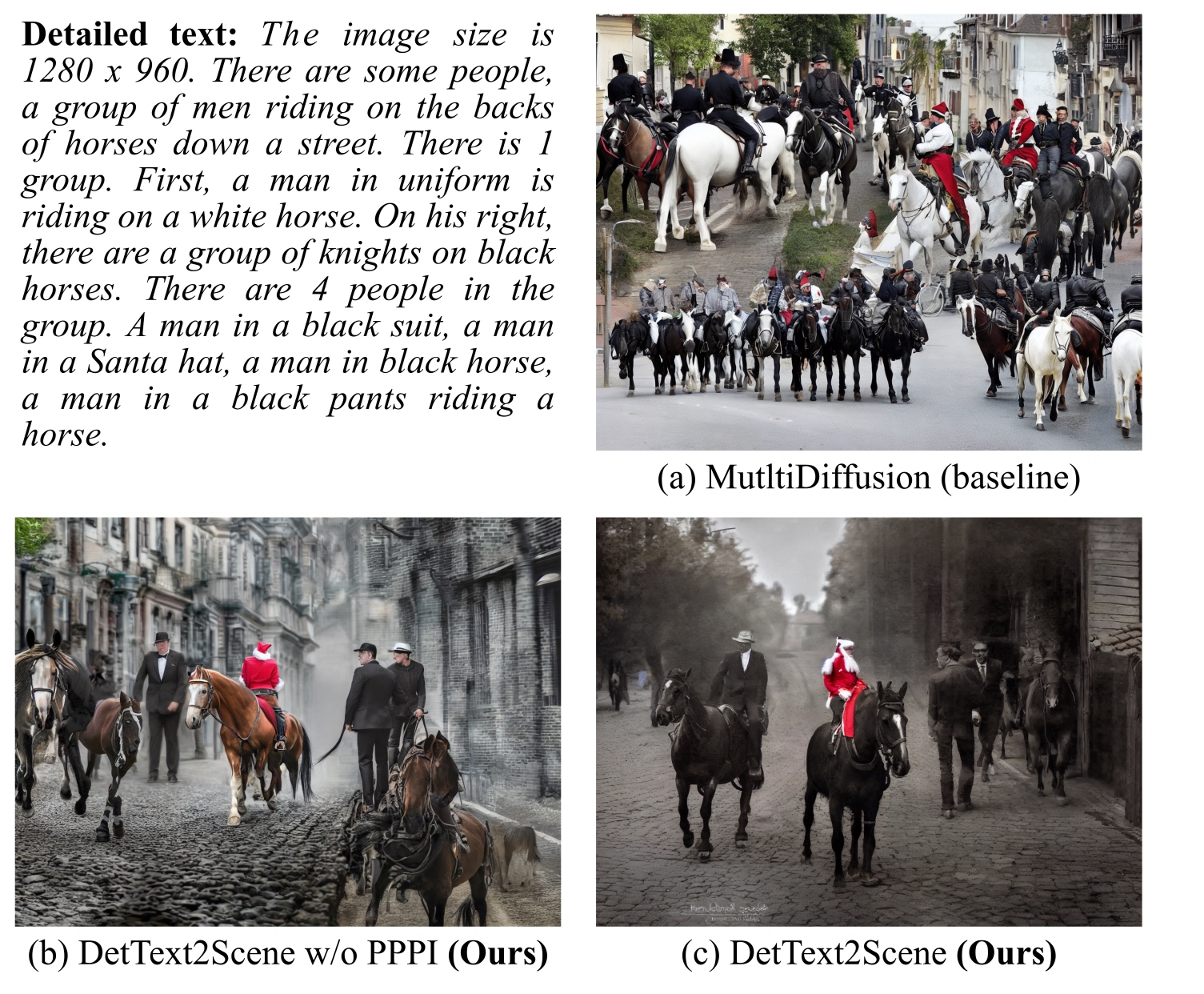}
    \vspace{-.7em}
    \caption{Ablation studies of our DetText2Scene. Compared to prior works, (a) MultiDiffusion~\citep{bar2023multidiffusion} and (b) our hierarchical keypoint-box layout generation and VCJD significantly enhanced \textit{faithfulness} and \textit{controllability}. And PPPI increases \textit{naturalness} as shown in (c). All the images were generated with the size of (1280$\times$960).}
    \label{fig6}
\end{figure}

\subsection{Large scene synthesis from spatial controls with detailed text}
\label{subsec_eval3}
We evaluate our large-scene synthesis method, which utilizes detailed text and input layout, against Region-based MultiDiffusion~\citep{bar2023multidiffusion}, the only published method known to us that generates large scenes using both spatial controls and text input with our knowledge.
To ensure a fair comparison, we meticulously selected 100 real-world complex scene images from the hierarchical keypoint-box layout dataset. These images have a total of 5 to 15 humans and objects and a size smaller than 1,500 pixels.

\paragraph{Qualitative comparisons.}
As illustrated in Fig.~\ref{fig5}, the results from Region-based MultiDiffusion given input layout with detailed text exhibit different scales for each instance, resulting in a disconnected and awkward scene. In contrast, our method successfully generates faithful results that appear natural in the global context.

\paragraph{Quantitative comparison.}
We utilize the global CLIP score \citep{hessel2021clipscore} to evaluate both faithfulness and  $N_{\text{human}}$ matching performance, assessing the controllability of the methods. Our method outperforms the baselines, achieving a higher CLIP score and F1 scores for $N_{\text{human}}$ matching, as demonstrated in Table~\ref{tab3}.

\paragraph{User study.}
Likewise, we conducted a user study employing the same criteria as the user study in Section~\ref{subsec_eval1}, with the addition of spatial coherence between the input layout and the location of generated instances.
As shown in Table~\ref{tab3}, out of the 8,040 responses from 201 participants, our method was significantly preferred, with 62.4\%-71.5\% of the votes in all four criteria.

\subsection{Ablation studies}
\label{subsec_ablation}
Our baseline model, MultiDiffusion, exhibits excessive duplication and disconnectedness of humans and objects, suggesting a lack of faithfulness, controllability, and naturalness, as shown in Fig.~\ref{fig6}(a). Our hierarchical keypoint-box layout generation and VCJD significantly enhance faithfulness and controllability, as represented in Fig.~\ref{fig6}(c). Additionally, PPPI increases naturalness without disconnectivity, as shown in Fig.~\ref{fig6}(d).
We provide quantitative results on ablation studies in the supplementary.

\section{Discussion}
\label{sec:Discussion}

\paragraph{Limitation.}
Although LLM shows promising results for generating a keypoint-box layout, the capability of understanding visual context may be limited.
For example, 3D information, such as depth information, is difficult to be addressed with LLM. 
The quality of generated large-scene results depends on the power of text-to-image diffusion models.
The limitation of the Stable diffusion 1.5~\citep{rombach2022high} includes falling short of achieving 1) complete photorealism, 2) compositionality, 3) proper face generation, 4) generating images with other languages except for English, and so on.

\section{Conclusion}
We propose DetText2Scene, a novel detailed human-centric text description-driven large-scale image synthesis with 1) a hierarchical keypoint-box layout conversion from the detailed text by LLM, 2) a view-wise conditioned joint diffusion process, and 
 3) a pixel perturbation-based pyramidal interpolation.
Our method significantly outperforms existing methods qualitatively and quantitatively with strong faithfulness, high controllability, and excellent naturalness in a global context for the given detailed text descriptions.
\section*{Acknowledgements}
This work was supported in part by the National Research Foundation of Korea(NRF) grants funded by the Korea government(MSIT) (NRF-2022R1A4A1030579, NRF-2022M3C1A309202211) and Creative-Pioneering Researchers Program through Seoul National University. Also, the authors acknowledged the financial support from the BK21 FOUR program of the Education and Research Program for Future ICT Pioneers, Seoul National University.

{
    \small
    \bibliographystyle{ieeenat_fullname}
    \bibliography{main}
}
\clearpage
\setcounter{page}{1}

\appendix

\noindent\large\textbf{Supplementary Material}
\normalsize

\setcounter{equation}{0}
\setcounter{figure}{0}
\setcounter{table}{0}
\setcounter{page}{1}
\makeatletter
\renewcommand{\theequation}{S\arabic{equation}}
\renewcommand{\thefigure}{S\arabic{figure}}
\renewcommand{\thetable}{S\arabic{table}}

\section{Implementation details}
\subsection{Hierarchical keypoint-box layout generation}
\paragraph{Details of dataset.}
To generate a keypoint-box layout from detailed human-centric text descriptions in large scene synthesis, we utilized group information and proposed a hierarchical generation process to address the challenges of directly generating multiple instances from detailed texts. The CrowdCaption dataset~\cite{wang2022happens} provide the group bounding box and group captions for  human-centric large scenes. It contains 11,161 images with 21,794 group region and 43,306 group captions. Each group has an average of 2 captions.

From the CrowdCaption dataset~\cite{wang2022happens}, we generated hierarchical keypoint-box layout dataset for instruction tuning of LLMs. Through (1) extraction of dense caption and visual location using existing models of dense captioning (GRiT~\citep{wu2022grit},ViT-B), image captioning (BLIP~\citep{li2022blip},ViT-B) and pose estimation (ViTPose-B~\citep{xu2022vitpose}) and (2) aligning process of acquired information based on IoU matching, we construct the hierarchical keypoint-box layout dataset which includes captions of image, groups and instances and visual information such as keypoint and bounding box of instances. Since each group region has multiple group captions, we augmented the instruction-answer pairs by matching several group captions in each image. Moreover, we consider the instances located in out of all groups for making more keypoint-box sample as non-group cases. The hierarchical keypoint-box layout dataset consists of 3 sets of instruction-answer pairs for instruction tuning of each LLMs. There are 41420, 41420 and 103329 instruction-answer pairs for Nat2Hier, global grounding and local grounding LLM, respectively. 
The averaged number of extracted instances per group region are 2.4 person (min: 1, max: 9) and 1.4 object (min: 0, max: 9). The example of instruction-answer pairs are visualized in the Fig.~\ref{fig_supp1} and Fig.~\ref{fig_supp2}.

\paragraph{Details of instruction tuning.}
Training details of our hierarchical keypoint-box layout generation are presented in the Table~\ref{table:implementation}. For instruction tuning, we utilized the open-source toolkit of training LLMs~\cite{gao2023llamaadapterv2,zhang2023llamaadapter} for single-turn and multi-turn models. For efficient usage of resources, the quantized parameter efficient fine-tuning is adopted. Moreover, we used 4 A100 GPUs and training time was about 5 hours, 5 hours and  1 day for Nat2Hier, global grounding and local grounding LLM, respectively. 

\begin{table*}[!h]
\centering
\small
\begin{tabular}{llllll}

\hline
                     & Model & Annotations              & Batch size & Epcoh & Learing rate    \\ \hline
Nat2Hier LLM & LLAMA2-7B             & Hierarchical structured descriptions                & 8          & 4     & 3e-5      \\                     
Global grounding LLM & LLAMA2-7B             & Group box                & 8          & 4     & 3e-5      \\
Local grounding LLM  & LLAMA2-7B             & Instance keypoint \& box & 4          & 4     & 3e-5            \\ \cline{1-6}   
\end{tabular}
\caption{Training details of LLMs for hierarical keypoint-box layout generation. The Nat2Hier LLM converts the detailed text into hierarchical structured text, and global grounding LLM estimates suitable locations of group boxes using this hierarchical structured text. Next, local grounding LLM estimates location of instances keypoint-box using output of Nat2Hier and global ground LLM. The Nat2Hier, global grounding and local grounding LLM refers to left, middle and right LLMs in stage 1 of the Fig. 2 in the main paper.}
\label{table:implementation}
\end{table*}

\subsection{View-wise conditioned joint diffusion}
\paragraph{Details of keypoint-based diffusion models.}

We adopt ControlNet~\citep{zhang2023adding} as our kepoint-based diffusion models.
ControlNet~\citep{zhang2023adding} is a neural network architecture that enhances diffusion models by incorporating additional conditioning information. It achieves this by duplicating the weights of neural network blocks into two copies: a "locked" copy and a "trainable" copy. The "trainable" copy is responsible for learning the conditioning information, while the "locked" copy maintains the original diffusion model's capabilities. This approach ensures that training with limited datasets of image pairs will not compromise the stability and performance of the pre-trained diffusion model.
ControlNet~\citep{zhang2023adding} utilizes a technique known as "zero convolution," which involves employing $1\times1$ convolutions with both weight and bias initialized to zeros. This approach enables training on small-scale or even personal devices, making ControlNet a versatile and accessible tool.

Among the diverse types of pre-trained ControlNet~\citep{zhang2023adding} models, we focus on keypoint-conditioned ControlNet. To train this model, learning-based pose estimation techniques, such as OpenPose~\citep{cao1812openpose}, are employed to identify humans in internet images. The extracted pose information is then paired with corresponding image captions, resulting in a dataset of 200,000 pose-image-caption pairs. 
For more details, please refer to \citep{zhang2023adding}.

\paragraph{Details of attention modulation.}
The attention map $A$ in Stable diffusion~\cite{rombach2022highresolution} which measures the relevance between different tokens in the text and image representations, is defined as follows:
\begin{align}
A=\textnormal{softmax}\left(\frac{Q K^{\top}}{\sqrt{d}}\right),
\end{align}
where $Q$ and $K$ represent the query and key vectors, respectively, and $d$ denotes the dimension of these vectors.

To enhance the model's ability to localize objects and keypoints in the generated images, we employ an attention modulation method inspired by \citep{kim2023dense}.
This method involves conditioning the attention score computation with additional inputs: a full-text description $y$ and a set of dense caption pairs $\{(y_n,s_n)\}_{n=0}^{N-1}$. Here, $y_n$ provides a non-overlapping description of each human or object in the image, $s_n$ represents a binary mask extracted from the corresponding bounding box or pose annotation. 

The attention modulation aims to strengthen the connections between text tokens and their corresponding image regions, while also encouraging interactions between tokens belonging to the same object. This is achieved by modifying the attention scores as follows:
\begin{align}
& A^{\prime}=\textnormal{softmax}\left(\frac{Q K^{\top}+M}{\sqrt{d}}\right), \nonumber \\
& M=\lambda_t \cdot R \odot M_{\textnormal{pos}} \odot(1-S) \\
& -\lambda_t \cdot(1-R) \odot M_{\textnormal{neg}} \odot(1-S), \nonumber
\end{align}
The query-key pair condition map $R$ determines whether to increase or decrease the attention score for a particular pair of tokens. Tokens belonging to the same segment form positive pairs, their attention scores being boosted, while tokens from different segments form negative pairs, their attention scores being suppressed.
The matrices $M_{\text{pos}}, M_{\text{neg}}$ are introduced to preserve the original value range of the attention scores, maintaining the model's pre-trained generation capabilities. The matrix $S$ reflects the relative size of each object, allowing for modulation adjustments based on object size.
Finally, a timestep-dependent modulation factor $\lambda_t$ is introduced to gradually reduce the degree of modulation as the diffusion process progresses, preventing excessive alterations that could degrade image quality.
$R, S, M_{\text{pos}}, M_{\text{neg}}$ are calculated from $y$ and $\{(y_n,s_n)\}_{n=0 }^{N-1}$. 
For more details, please refer to \citep{kim2023dense}.

\subsection{Pixel perturbation-pyramidal interpolation}
\paragraph{Pyramidal image generation.}
 In pyramidal image generation, we proceed the reverse process $P$ times. After each stage of generation, we expanded the height and the width of image with pixel perturbation interpolation. At final stage, pixel perturbation interpolation is not applied. We set $P$ as 3 for qualitative evaluation and 2 for qualitative evaluation. Furthermore, except for first stage of pyramidal image generation, we set $T_{p}=0.5$ within normalized timestep $t \in [0, 1]$ for all the phases $p \in \{1, 2\dots,P-1\}$.

 \paragraph{Pixel perturbation-based interpolation.}
 Pixel perturbation swaps each pixel in the interpolated image with the existing pixel in $d_{\textnormal{pert}}$ with a certain probability $\gamma_{\textnormal{pert}}$. At this time, all experiments were performed using Lanczos interpolation, and $\gamma_{\textnormal{pert}}=0.05$. If $\gamma_{\textnormal{pert}}$ is large, the high-frequency component of the interpolated image will be increased and the result after going through the diffusion step will also emphasize high-frequency. If $\gamma_{\textnormal{pert}}$ is small, an image in which the low-frequency component is dominant can be obtained. If $\gamma_{\textnormal{pert}}$ exceeds a certain level, there is a point where semantic information is lost due to excessive pixel swapping, and artifacts occur. Additionally, the larger $d_{\textnormal{pert}}$ is, the wider the pixels are swapped, which results in losing more semantic information and losing the original purpose. Therefore, $d_{\textnormal{pert}}$ was decided to 1. Further experiments on the hyperparameters are represented in Section~\ref{secc4:study_PPPI}.

\section{Experimental details}


\subsection{Details on evaluation of large image generation from detailed text}

\paragraph{Quantitative comparison.}
We calculated global CLIP score~\citep{hessel2021clipscore} to measure the faithfulness and the numerical matching performance of the number of human instances ($N_{\text{human}}$ matching) between the text prompt and the generated images to assess the controllablity of the methods. 
\textbf{(1) Global CLIP score:} We calculate the cosine similarity between generated image and corresponding text prompt including global and group description using CLIP-ViT-B/32 model.
\textbf{(2) $N_{\text{human}}$ matching:} 
To evaluate the controllability of our proposed DetText2Scene, We measured the numerical matching score and reported the precision, recall and F1 score by following ~\citep{feng2023layoutgpt}. We compared the number of human between ground truth from input description and the human counting number of generated image estimated by YOLOv7~\citep{wang2023yolov7}.

\begin{figure}[h]
    \centering
    \includegraphics[width=0.45\textwidth]{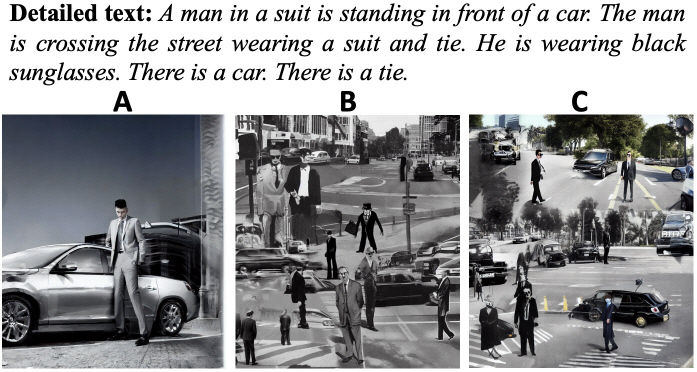}
    \caption{An example of the questionnaires for evaluation of large image generation from detailed text. The users were asked to rank the given three images in order of faithfulness, naturalness from global context, and naturalness from physical structure perspective. The order of the presented images were randomly given.}
    \label{fig_supp_comp}
\end{figure}

\paragraph{User study.}
We conducted a user study to further evaluate the faithfulness and naturalness using a crowd sourcing.
Participants were presented with large-scene images generated by  MultiDiffusion, SyncDiffusion and our DetText2Scene methods.
They were then asked to rank the methods with following the guidelines: Rank the images in order of (1) their faithfulness with the text without missing objects and incorrect binding between words and objects, (2) their naturalness from global context, and (3) their naturalness from a physical structure perspective.
The order of the images was shuffled.
We crafted detailed caption from random 5 CrowdCaption test images.
We placed the results from 3 methods side-by-side.
A total of 201 people completed the survey, providing 9,045 votes.
Here, we present some examples of the questionnaires of user study comparing MultiDiffusion, SyncDiffusion, and our DetText2Scene in Fig.~\ref{fig_supp_comp}.

\subsection{Details on evaluation of keypoint-box layout generation part}

As our method is the first work for predicting the keypoint and box layouts simultaneously from the detailed description, we thoroughly evaluated our method on the 100 real-world complex scene images from the hierarchical keypoint-box layout dataset. The total number of humans and objects in each image is from 5 to 15 and a size of image is smaller than 1,500 pixels.
\paragraph{Quantitative comparisons.}
To evaluate the controllability and quality of generated keypoint-box layout from hierarchical detailed description, we measured performance of numerical matching and spatial matching of visual location of group and instance. For numerical matching, we checked that the desired number of group, human and object matches the generated keypoint-box layout and reported precision, recall and F1 score as used in LayoutGPT~\cite{feng2023layoutgpt}. For spatial matching, we checked location matching between the desired group box location and the center of the generated group box layout in two spatial cases (\textit{left, right}). We added the spatial condition of group box to input prompt. Moreover, we measured the performance of group box inclusion, checking that the generated keypoint is located in the corresponding group box.

\subsection{Details on evaluation of large image generation part}
\label{B3_largeimagegen}

\begin{figure}[!t]
    \centering
    \includegraphics[width=0.45\textwidth]{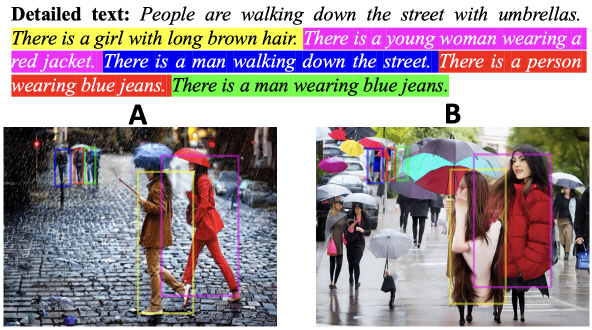}
    \caption{An example of the questionnaires for evaluation of large image generation from spatial controls with detailed text. The users were asked to rank the given two images in order of faithfulness, spatial coherence, naturalness from global context, and naturalness from physical structure perspective. The order of the presented images were randomly given.}
    \label{fig_supp_region}
\end{figure}

\paragraph{User study.}
We conducted a user study to further evaluate the faithfulness and naturalness using a crowd sourcing.
Participants were presented with large-scene images generated by region-based MultiDiffusion~\citep{bar2023multidiffusion} and our DetText2Scene methods.
They were then asked to rank the methods with the same guidelines as used in above.
We crafted detailed caption from random 5 CrowdCaption test images.
We placed the results from 2 methods side-by-side.
A total of 201 people completed the survey, providing 8,040 votes.
Here, we present some examples of the questionnaires of user study comparing region-based MultiDiffusion and our DetText2Scene in Fig.~\ref{fig_supp_region}.

\begin{figure*}[t]
    \centering
    \includegraphics[width=0.85\textwidth]{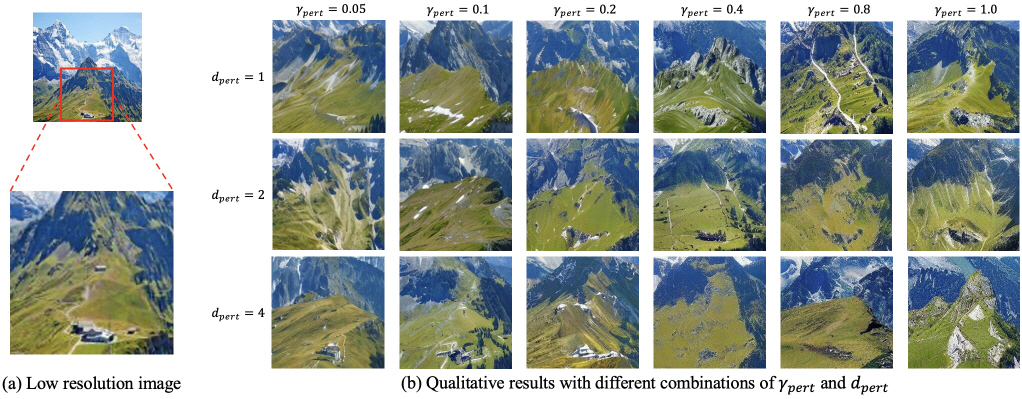}
    \caption{Results of generated images from low-resolution images with different $\gamma_{\textnormal{pert}}$ and $d_{\textnormal{pert}}$. As the $\gamma_{\textnormal{pert}}$ increases, the high frequency of the images increases, and as $d_{\textnormal{pert}}$ increases, the more structural changes occurs compared to the original image. All the images were generated with $T_p=0.5$, and text \textit{``the photo of the Alps"}.}
    \label{fig_pppi_param}
\end{figure*}

\begin{table*}[!htp]
          \centering
          \captionsetup[subtable]{position = below}
         \captionsetup[table]{position=top}
          \vspace{-1em}
          \begin{subtable}{0.26\linewidth}
              \centering
               \begin{adjustbox}{width=\linewidth}
\begin{tabular}{cccc}\toprule
\multirow{2}{*}{} &\multicolumn{3}{c}{Numerical matching} \\
&Prec.↑ &Rec.↑ &F1 ↑ \\\midrule
$N_{\text{group}}$ &1.000 &1.000 &1.000 \\
$N_{\text{human}}$ &1.000 &0.981 &0.991 \\
$N_{\text{obj}}$ &1.000 &0.911 &0.954 \\
\bottomrule
\end{tabular}
               \end{adjustbox}
          \end{subtable}%
          \hspace*{1em}
          \begin{subtable}{0.22\linewidth}
              \centering
               \begin{adjustbox}{width=\linewidth}
\begin{tabular}{ccc}\toprule
\multirow{2}{*}{} &\multirow{2}{*}{\makecell{Spatial matching \\  Acc.↑}} \\
& \\\midrule
Location &0.964 \\
Inclusion &0.800 \\
\bottomrule
\end{tabular}
               \end{adjustbox}
          \end{subtable}
          \caption{Quantitative evaluation of keypoint-box layout generation part. Our hierarchical keypoint-box layout generation using pre-trained LLMs demonstrates high performance quantitatively in terms of the numerical and spatial controllability. These are crucial for large scene synthesis from detailed text.}\label{tab_3_llm_quanti}
      \end{table*}

\begin{table*}[!htp]
           \centering
           \captionsetup[subtable]{position = below}
          \captionsetup[table]{position=top}
           \vspace{-.5em}
           \begin{subtable}{0.4\linewidth}
               \centering
                \begin{adjustbox}{width=\linewidth}
\begin{tabular}{ccccc}\toprule
\multirow{2}{*}{\textbf{\makecell{Quantitative \\ result}}} &\multirow{2}{*}{CLIP score↑} &\multicolumn{3}{c}{$N_{\text{human}}$ matching} \\\cmidrule{3-5}
& &Prec.↑ &Rec↑ &F1↑ \\\midrule
MultiDiffusion (Baseline) &17.400 &0.233 &0.993 &0.378 \\
DetText2Scene w/o PPPI (\textbf{Ours}) &30.666 &0.729 &0.931 &0.818 \\
DetText2Scene w/ PPPI (\textbf{Ours}) &32.097 &0.848 &0.933 &0.889 \\
\bottomrule
\end{tabular}
                \end{adjustbox}
           \end{subtable}%
           \hspace*{2em}
           \begin{subtable}{0.38\linewidth}
               \centering
                \begin{adjustbox}{width=\linewidth}
\begin{tabular}{ccccc}\toprule
\multirow{3}{*}{\textbf{User study}} &\multicolumn{3}{c}{Preference of ours $\uparrow$ } \\\cmidrule{2-4}
&\multirow{2}{*}{Faithfulness} &\multicolumn{2}{c}{Naturalness} \\
& &Global &Human \\\midrule
vs DetText2Scene w/o PPPI (\textbf{Ours}) &59.8\% &67.2\% &68.6\% \\
\bottomrule
\end{tabular}

                \end{adjustbox}
           \end{subtable}
\vspace{-0.5em}
           \caption{{Ablation studies of our DetText2Scene. Our hierarchical keypoint-box layout generation and VCJD quantitatively outperform the MultiDiffusion~\cite{bar2023multidiffusion} in terms of faithfulness and controllability. Moreover, PPPI increases the overall performance in multiple metrics and enhanced naturalness in user study.}}\label{tab3_ablation}
       \end{table*}


\section{Additional results}

\subsection{Additional results of large image generation from detailed text}
We present additional results of large image generation from detailed text with different seeds in Fig.~\ref{fig_supp_qual1}, Fig.~\ref{fig_supp_qual2}, Fig.~\ref{fig_supp_qual3}, Fig.~\ref{fig_supp_qual4}, Fig.~\ref{fig_supp_qual5}, Fig.~\ref{fig_supp_qual6}, and Fig.~\ref{fig_supp_qual7}. Moreover, we present large-scaled images from main paper in Fig.~\ref{fig_supp_qual8}, Fig.~\ref{fig_supp_qual9}, and Fig.~\ref{fig_supp_qual11}. Lastly, we demonstrate the controllability of DetText2Scene with slightly different text inputs in Fig.~\ref{fig_supp_qual12} and Fig.~\ref{fig_supp_qual13}.

\subsection{Additional results of keypoint-box layout generation part}

\paragraph{Quantitative results.}
As shown in the left of Table~\ref{tab_3_llm_quanti}, we measured the numerical matching performance for the number of group, human, and object instances ($N_{\text{group}}$, $N_{\text{human}}$ and $N_{\text{obj}}$ matching) between the text prompt and the generated keypoint-box layout. We achieved a high numerical matching performance over 0.91 in all cases and metrics.
As shown in the right of the Table~\ref{tab_3_llm_quanti}, we demonstrated the credible spatial matching performance of keypoint-box layout generation, measuring the accuracy of group location and instances inclusion within group box.

\paragraph{Qualitative results.}
We present qualitative results of keypoint-box layout generation part from our method in Fig.~\ref{fig_supp3} and Fig.~\ref{fig_supp8}.

\subsection{Additional results of large image generation part}
We present additional results of large image generation part from our method which is generated with text extracted from CrowdCaption dataset in Fig.~\ref{fig_supp9} and Fig.~\ref{fig_supp10}.

\subsection{Additional studies on PPPI}
\label{secc4:study_PPPI}
\paragraph{Qualitative results.}
We have two hyperparameters $\gamma_{\textnormal{pert}}$ and $d_{\textnormal{pert}}$ for PPPI. The  $\gamma_{\textnormal{pert}}$ is the probability of swapping two adjacent pixels, and the $d_{\textnormal{pert}}$ is the distance between two pixels that are recognized as adjacent pixels. As shown in Fig.~\ref{fig_pppi_param}, the higher $\gamma_{\textnormal{pert}}$  is, the more high frequencies are present in the generated image and the higher $d_{\textnormal{pert}}$ is, the more structural changes occur. In this experiment, we conducted ablation study by changing two hyperparameters of pixel perturbation interpolation. From randomly selected low-resolution image, we applied pixel perturbation interpolation with two hyperparameters, followed by a forward process with $T_p=0.5$, and then a reverse process.

\subsection{Ablation studies}
\paragraph{Quantitative results.} We conducted ablation studies of our DetText2Scene, measuring global CLIP score and the accuracy of the number of human instances($N_{\text{human}}$ matching) between the text prompt and the generated images. We reported the precision, recall and F1 scores for $N_{\text{human}}$ matching. As shown in the Table~\ref{tab3_ablation}, DetText2Scene w/o PPPI achieves a higher performance than MultiDiffusion~\cite{bar2023multidiffusion}. Then, DetText2Scene w/ PPPI demonstrated the best performance in terms of the faithfulness in CLIP score and the controllability in $N_{\text{human}}$ matching. 

\begin{figure}[!t]
    \centering
    \includegraphics[width=0.45\textwidth]{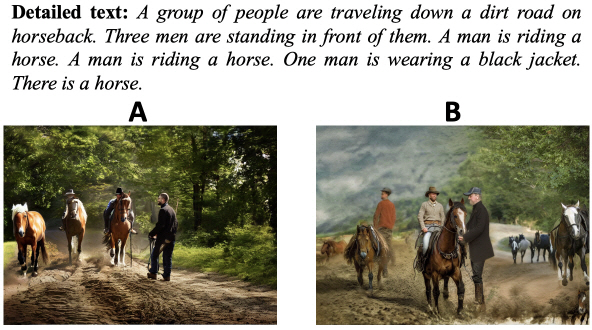}
    \caption{An example of the questionnaires for ablation study of our DetText2Scene. The users were asked to rank the given two images in order of faithfulness, naturalness from global context, and naturalness from physical structure perspective. The order of the presented images were randomly given.}
    \label{fig_supp_user_pppi}
\end{figure}

\begin{figure*}[!ht]
    \centering
    \includegraphics[width=0.96\textwidth]{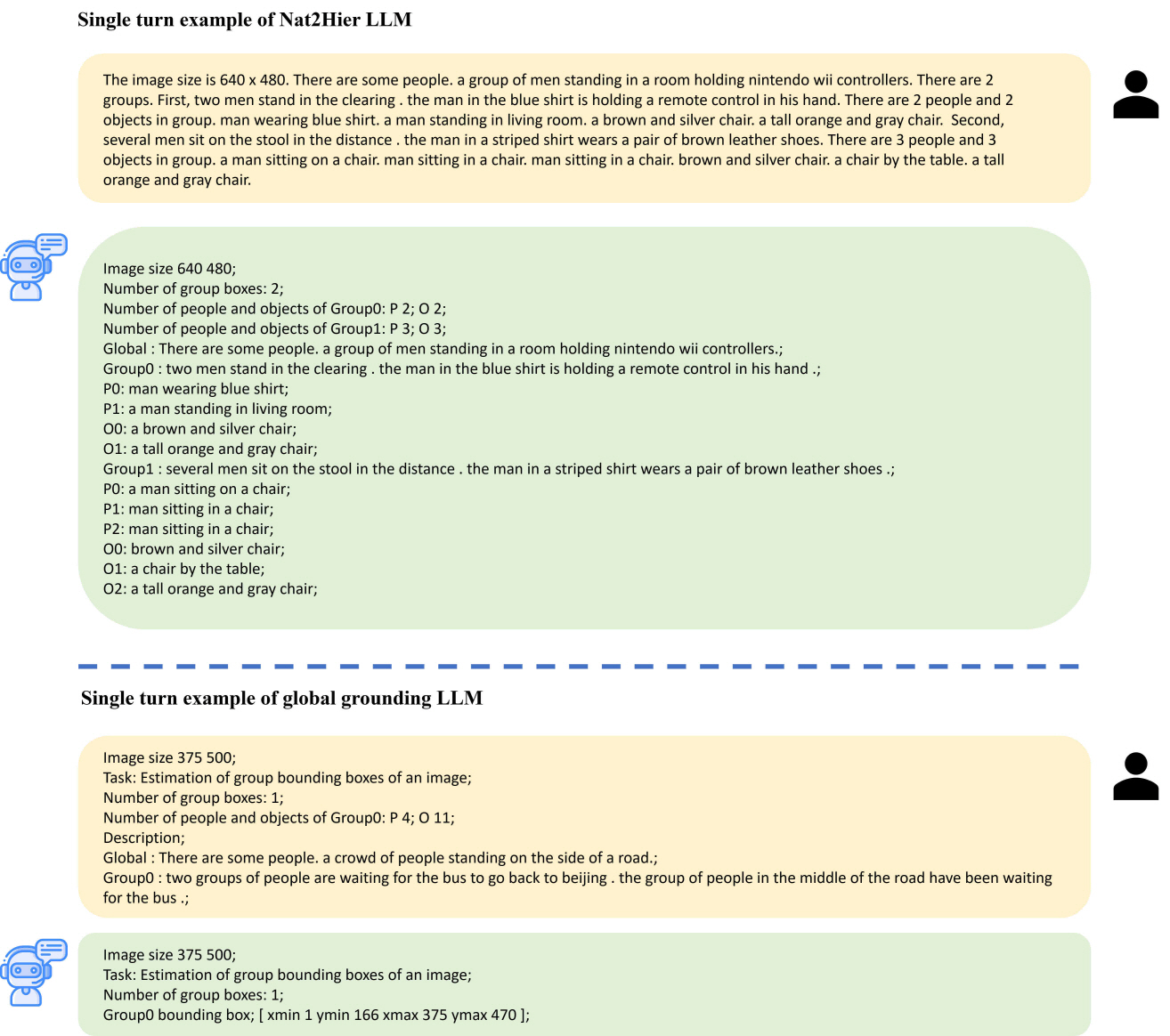}
    \caption{Examples of single turn dialog of Nat2Hier and global grounding LLM from the hierarchical keypoint-box layout dataset. These examples are instruction-answer pairs for instruction tuning of the Nat2Hier and global grounding LLMs. In single turn instruction-answer pairs for instruction tuning, we visualized the instruction in a yellow box and the answer(ground truth) in a green box, respectively. }
    \label{fig_supp1}
\end{figure*}

\begin{figure*}[t]
    \centering
    \includegraphics[width=0.96\textwidth]{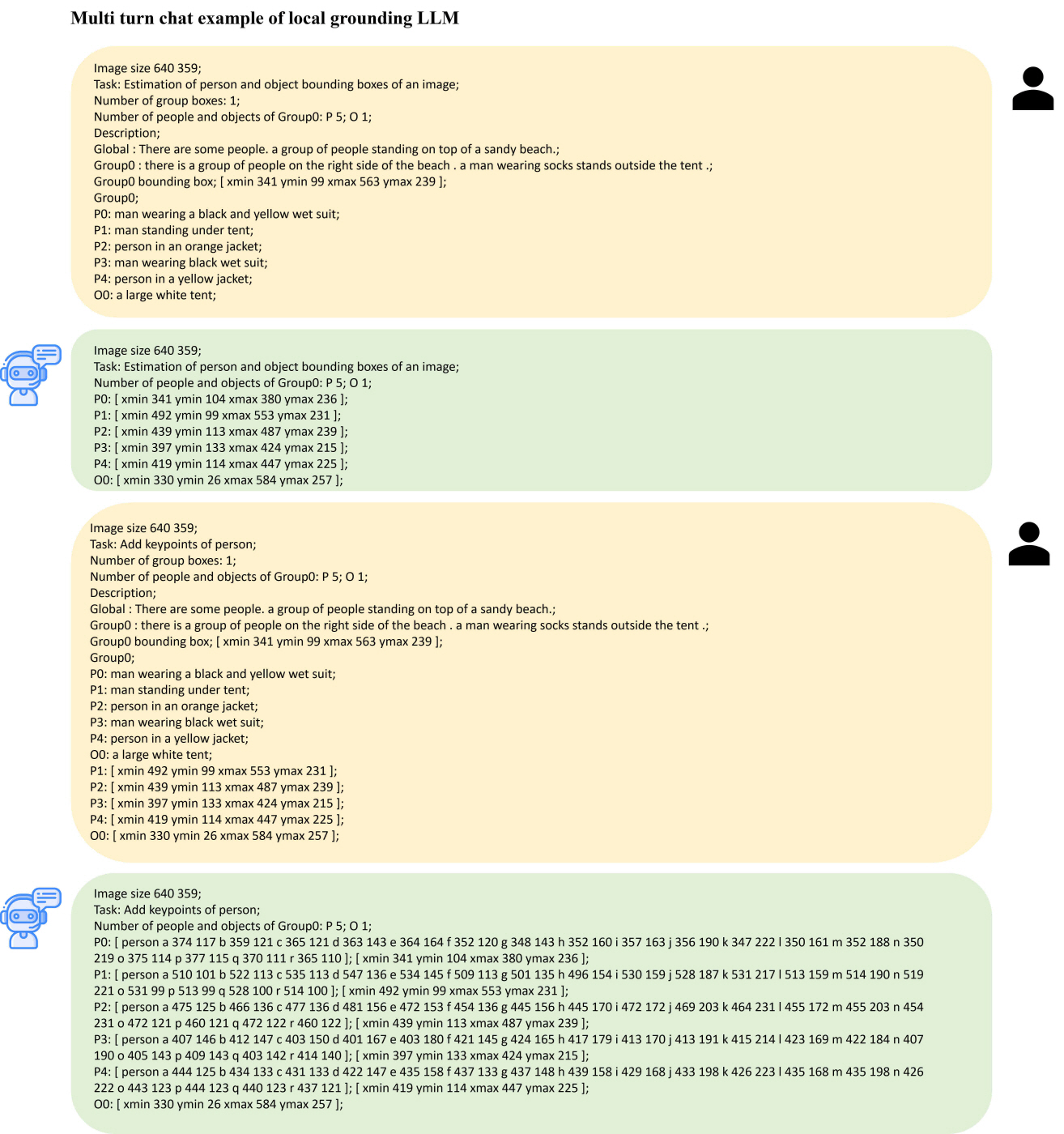}
    \caption{Examples of multi turn chat dialog of local grounding LLM from the hierarchical keypoint-box layout dataset. This example is instruction-answer pair for instruction tuning of the local grounding LLM. In multi turn instruction-answer pairs for instruction tuning, we visualized the instruction in a yellow box and the answer(ground truth) in a green box, respectively.}
    \label{fig_supp2}
\end{figure*}

\begin{figure*}[t]
    \centering
    \includegraphics[width=0.96\textwidth]{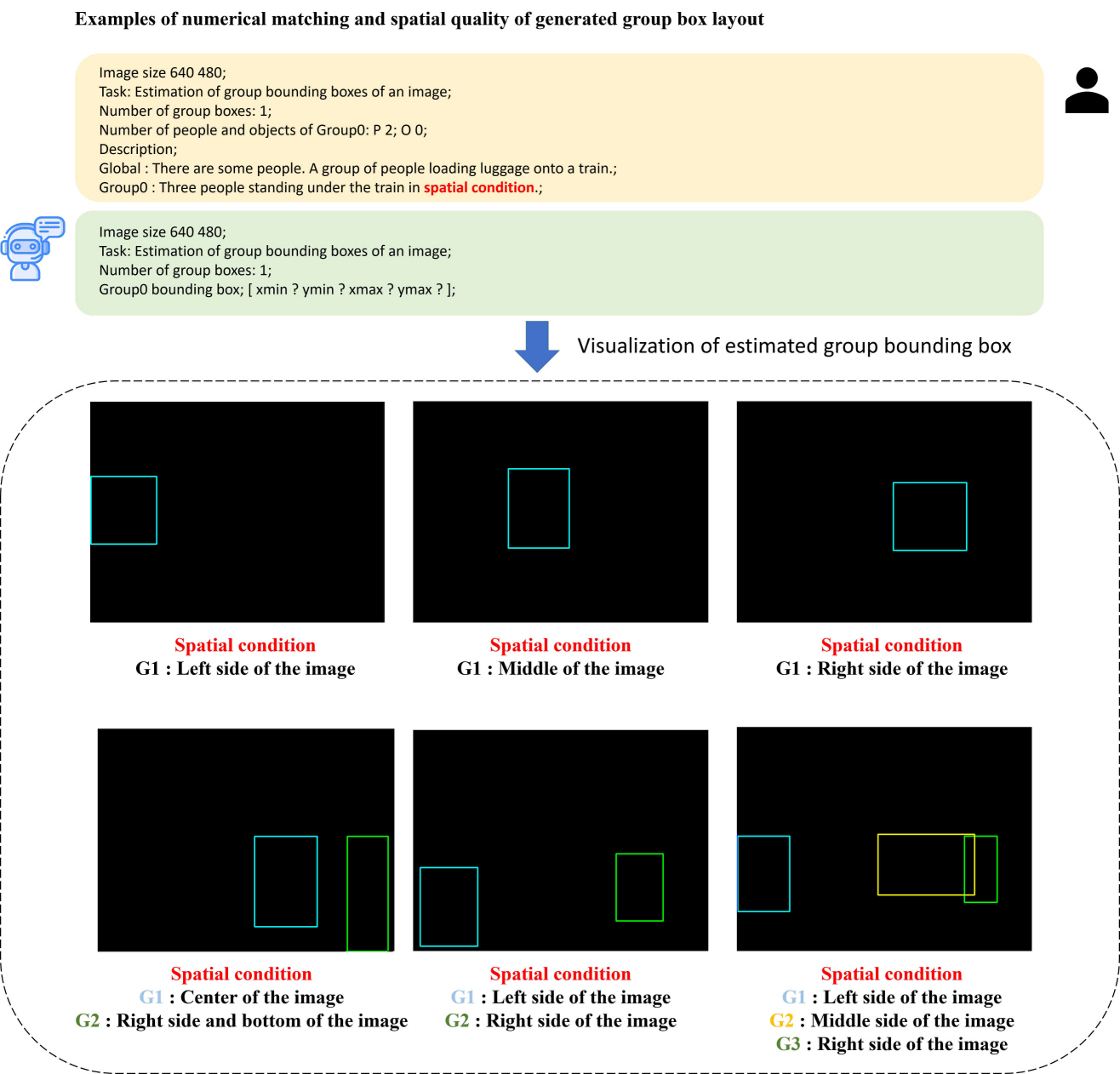}
    \caption{Examples of generated group box layout of global grounding LLM in diverse numerical and spatial cases. The example of single turn dialog is visualized and we varied the number of groups and spatial conditions. From the instruction containing spatial condition in yellow box, we estimate the group bounding box using global grounding LLM, like description in green box. We visualized the results of generated group box layout with different numerical and spatial conditions. The estimated group boxes from global grounding LLM are well-matched with numerical and spatial condition.}
    \label{fig_supp3}
\end{figure*}

\begin{figure*}[t]
    \centering
    \includegraphics[width=0.96\textwidth]{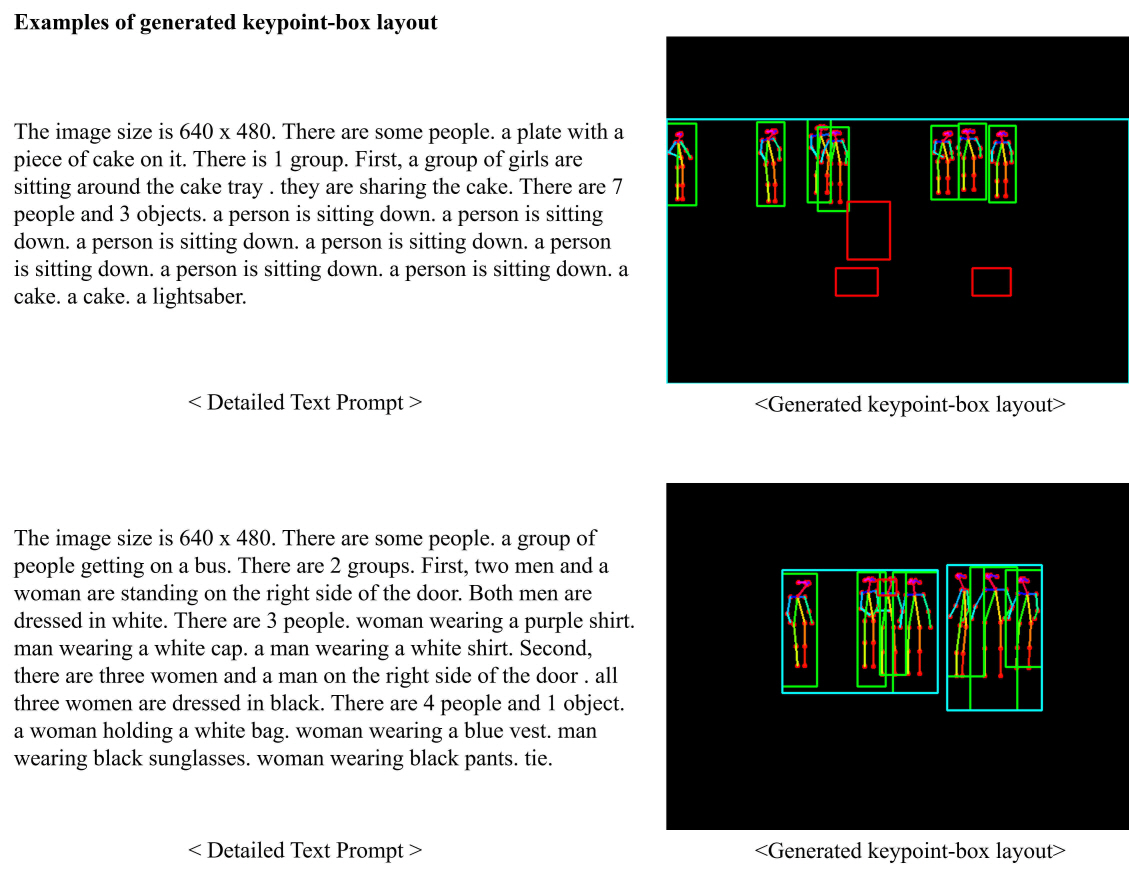}
    \caption{Examples of generated keypoint-box layout from the detailed text prompt. We visualized the keypoint and bounding box of instances and groups. The bounding boxes colored cyan, green and red represent the group, person and object box, respectively. These examples show that the estimated keypoint-box layout of instances is well-matched with numerical and spatial condition of detailed text prompt.}
    \label{fig_supp8}
\end{figure*}

\begin{figure*}[t]
    \centering
    \includegraphics[width=0.9\textwidth]{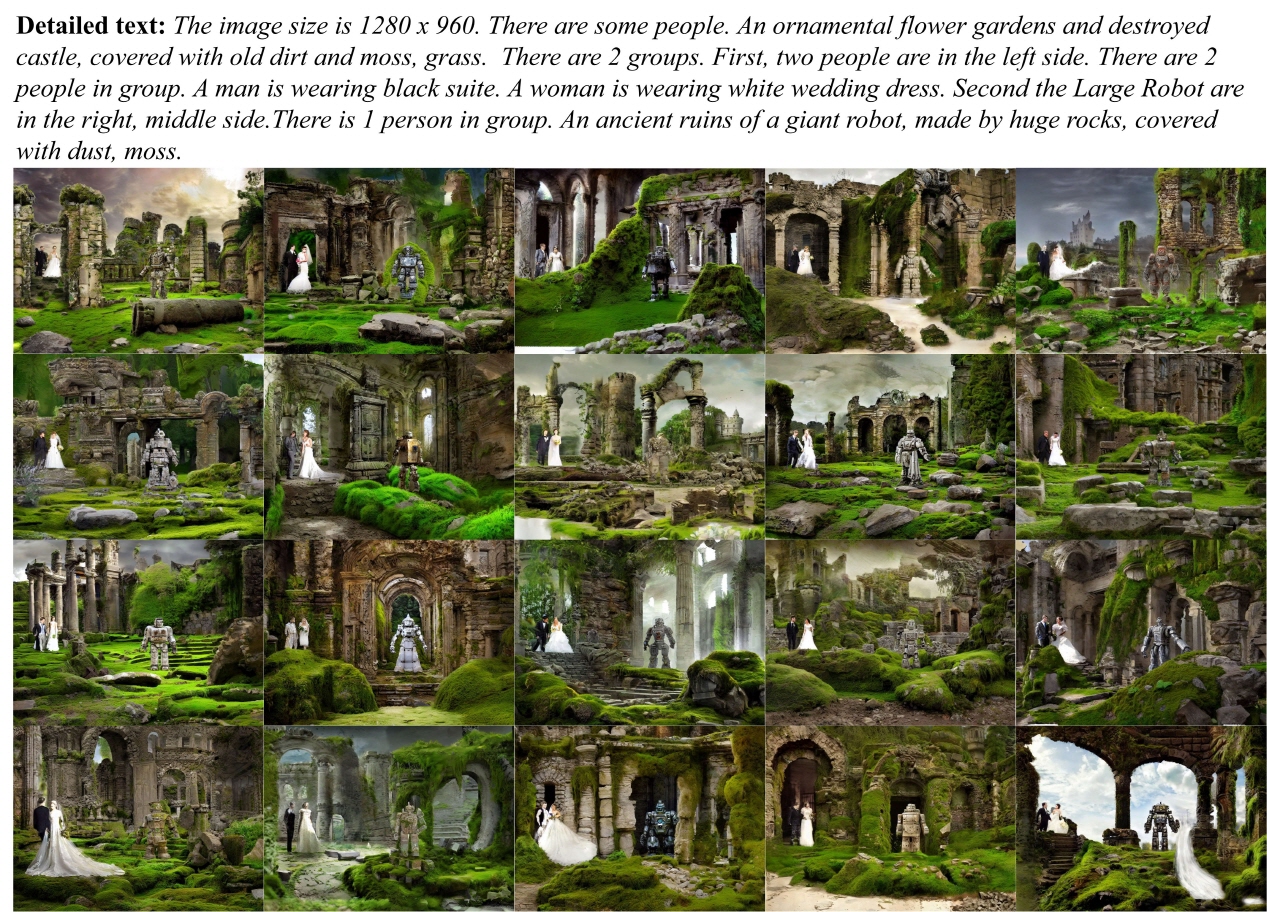}
    \includegraphics[width=0.9\textwidth]{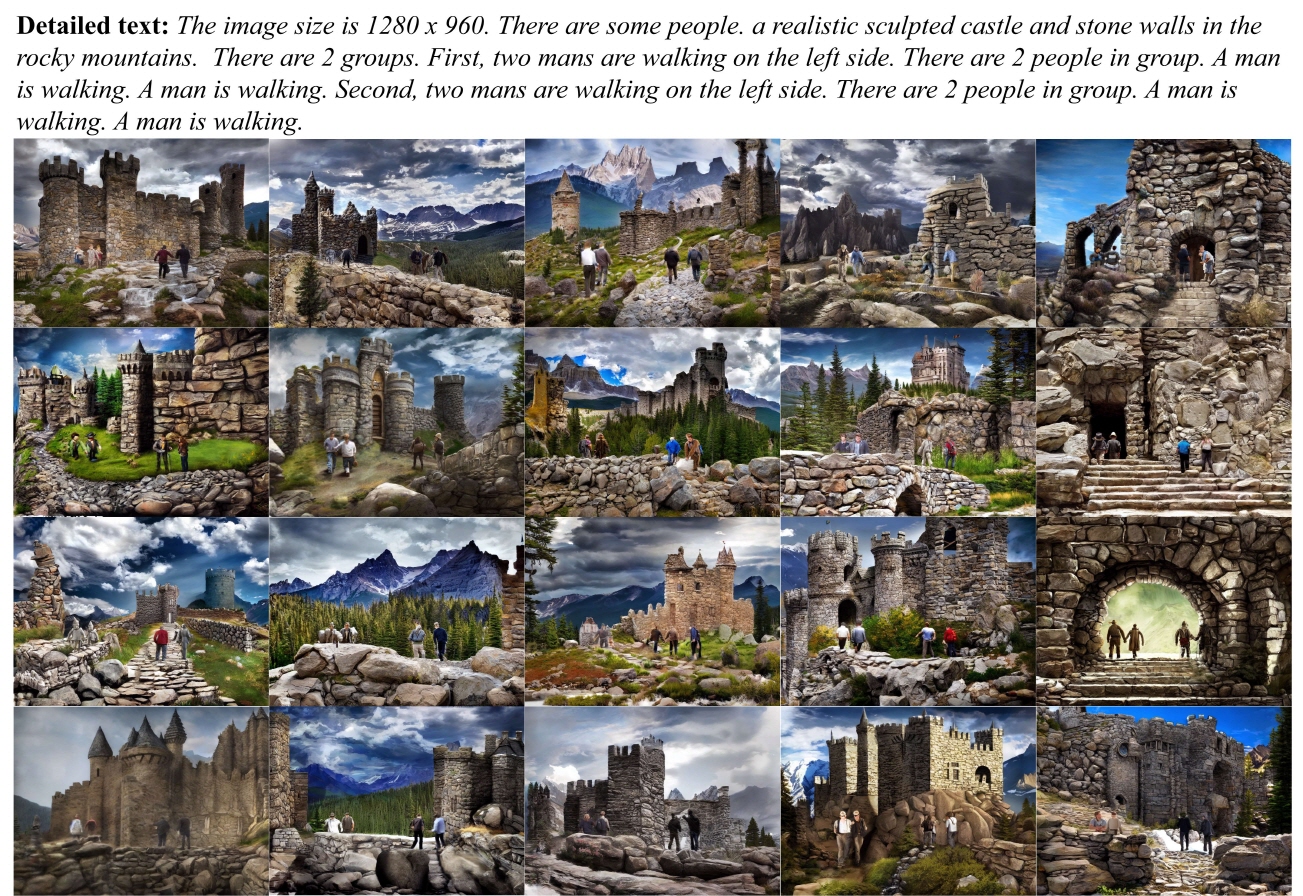}

    \caption{Qualitative results of our DetText2Scene with each detailed text and different seeds(1280$\times$960). Our DetText2Scene generates the diverse large scenes with different seeds while maintaining the high \textit{faithfulness}, \textit{controllability} and \textit{naturalness}.}
    \label{fig_supp_qual1}
\end{figure*}

\begin{figure*}[t]
    \centering
    \includegraphics[width=0.9\textwidth]{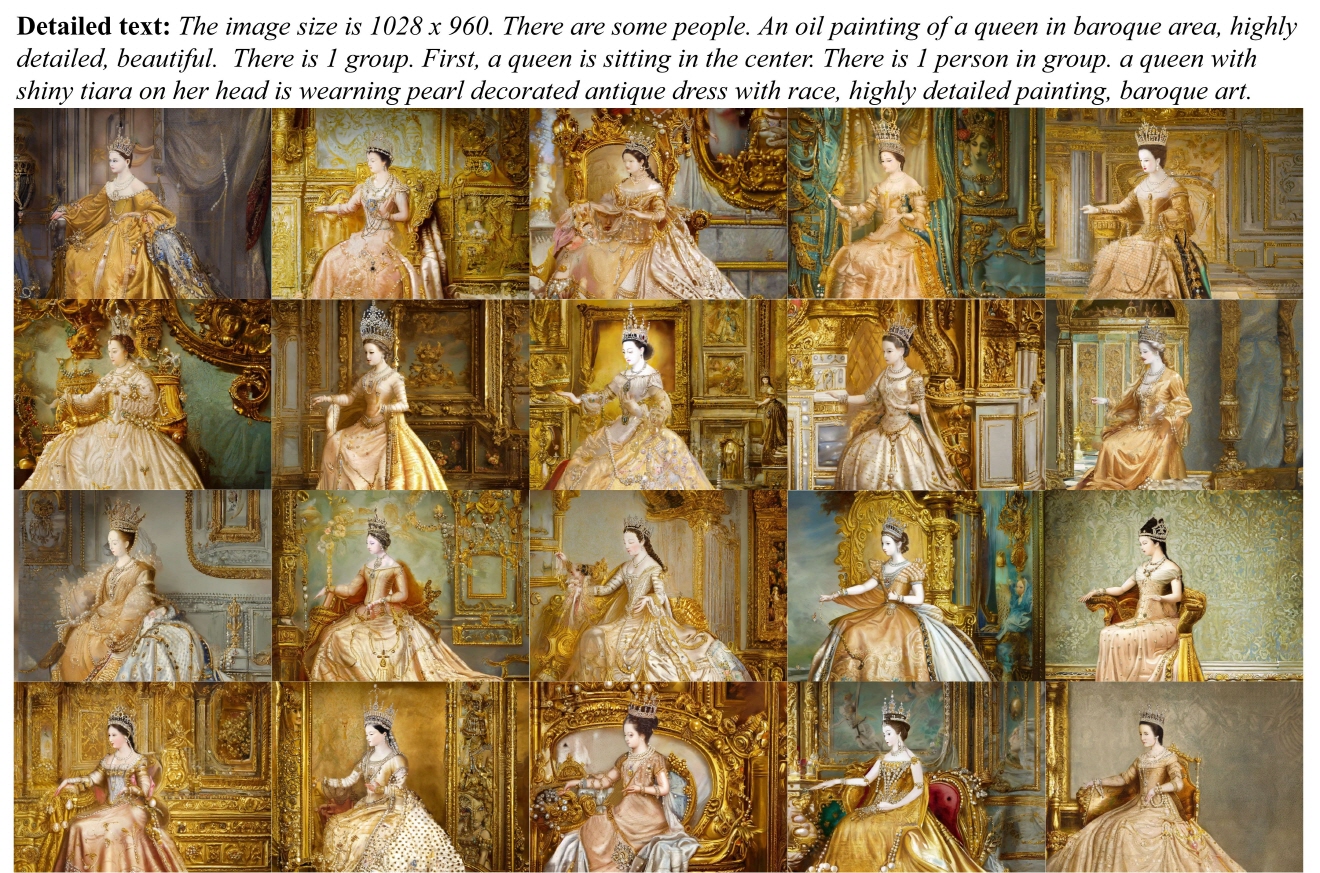}
    \includegraphics[width=0.9\textwidth]{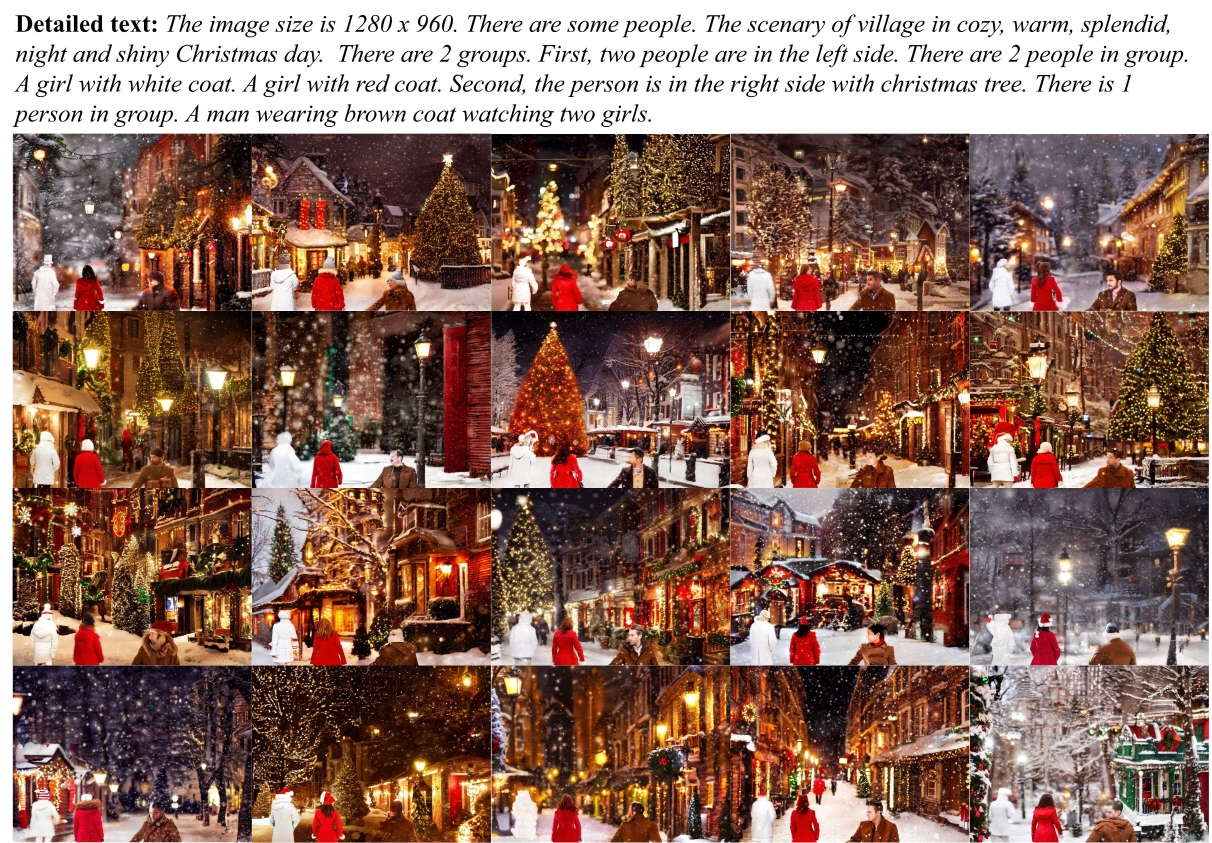}
    \caption{Qualitative results of our DetText2Scene with each detailed text and different seeds(1280$\times$960). Our DetText2Scene generates the diverse large scenes with different seeds while maintaining the high \textit{faithfulness}, \textit{controllability} and \textit{naturalness}.}
    \label{fig_supp_qual2}
\end{figure*}

\begin{figure*}[t]
    \centering
    \includegraphics[width=0.9\textwidth]{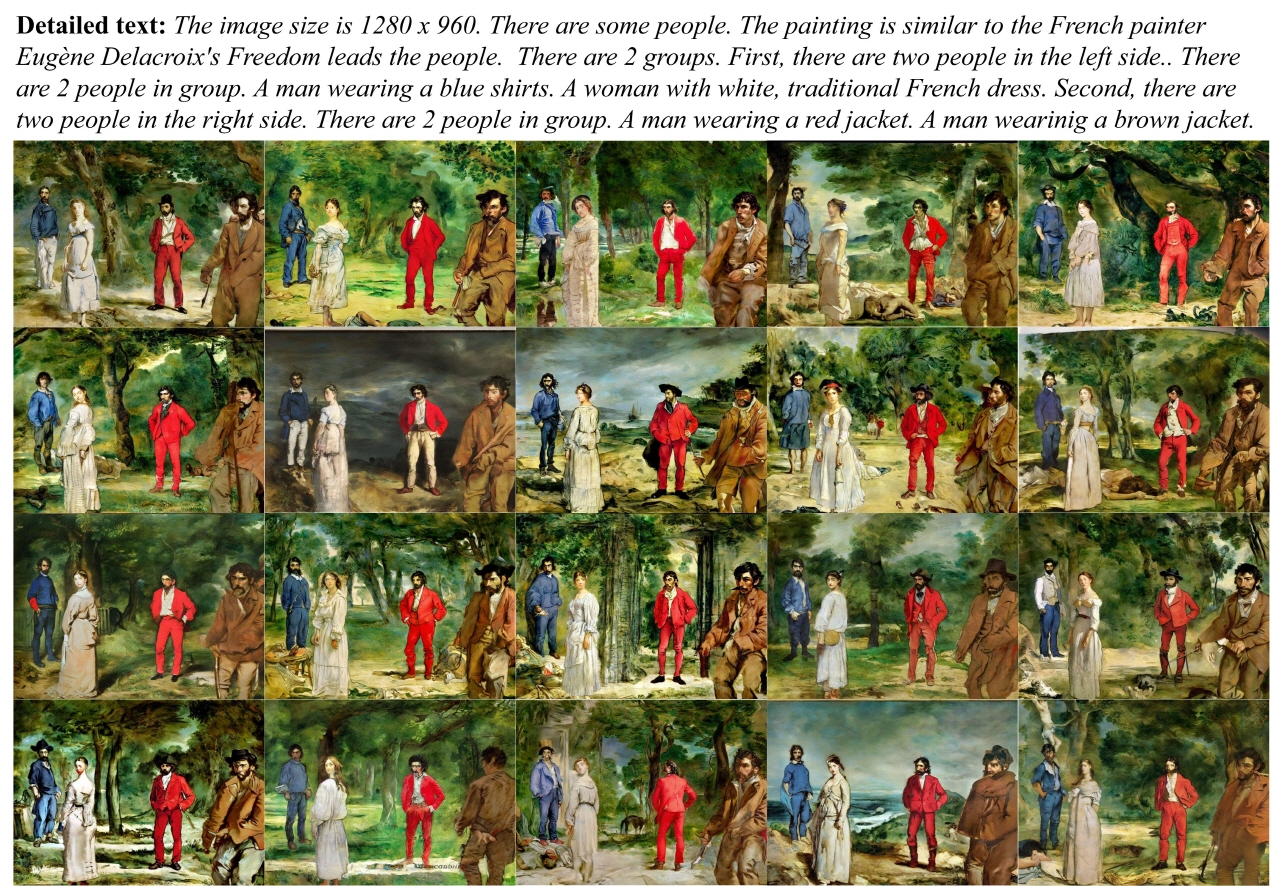}
    \includegraphics[width=0.9\textwidth]{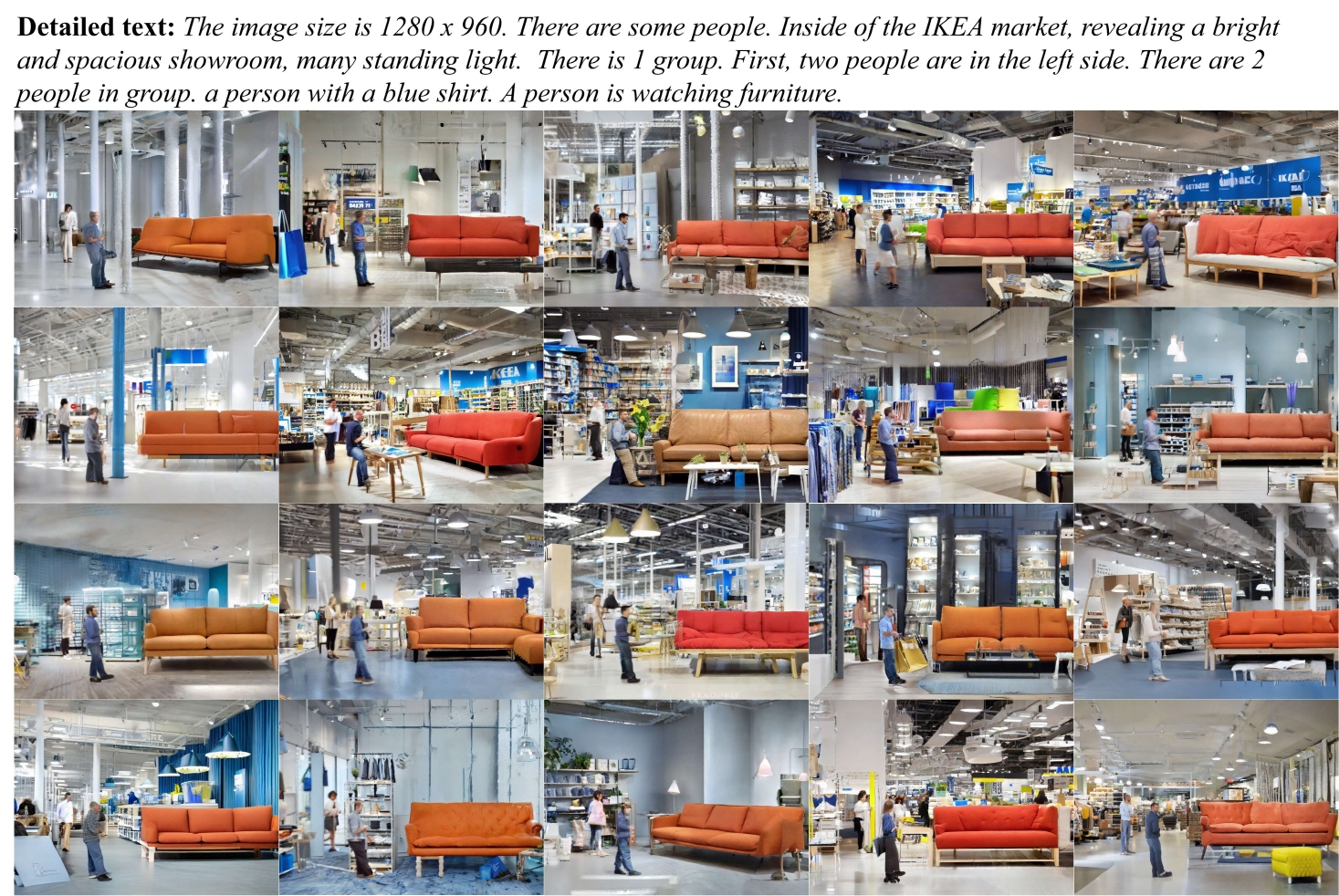}
    \caption{Qualitative results of our DetText2Scene with each detailed text and different seeds(1280$\times$960). Our DetText2Scene generates the diverse large scenes with different seeds while maintaining the high \textit{faithfulness}, \textit{controllability} and \textit{naturalness}.}
    \label{fig_supp_qual3}
\end{figure*}

\begin{figure*}[t]
    \centering
    \includegraphics[width=0.9\textwidth]{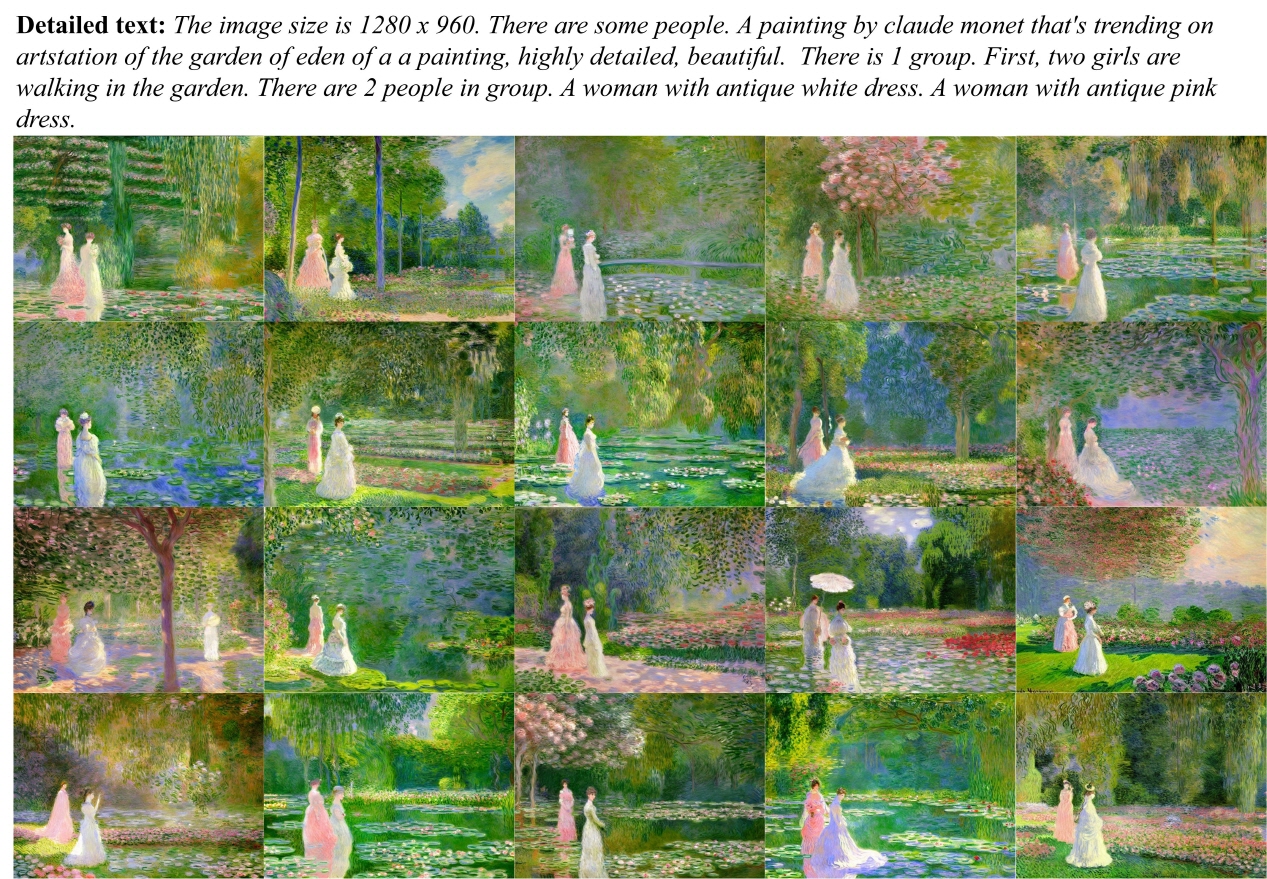}
    \includegraphics[width=0.9\textwidth]{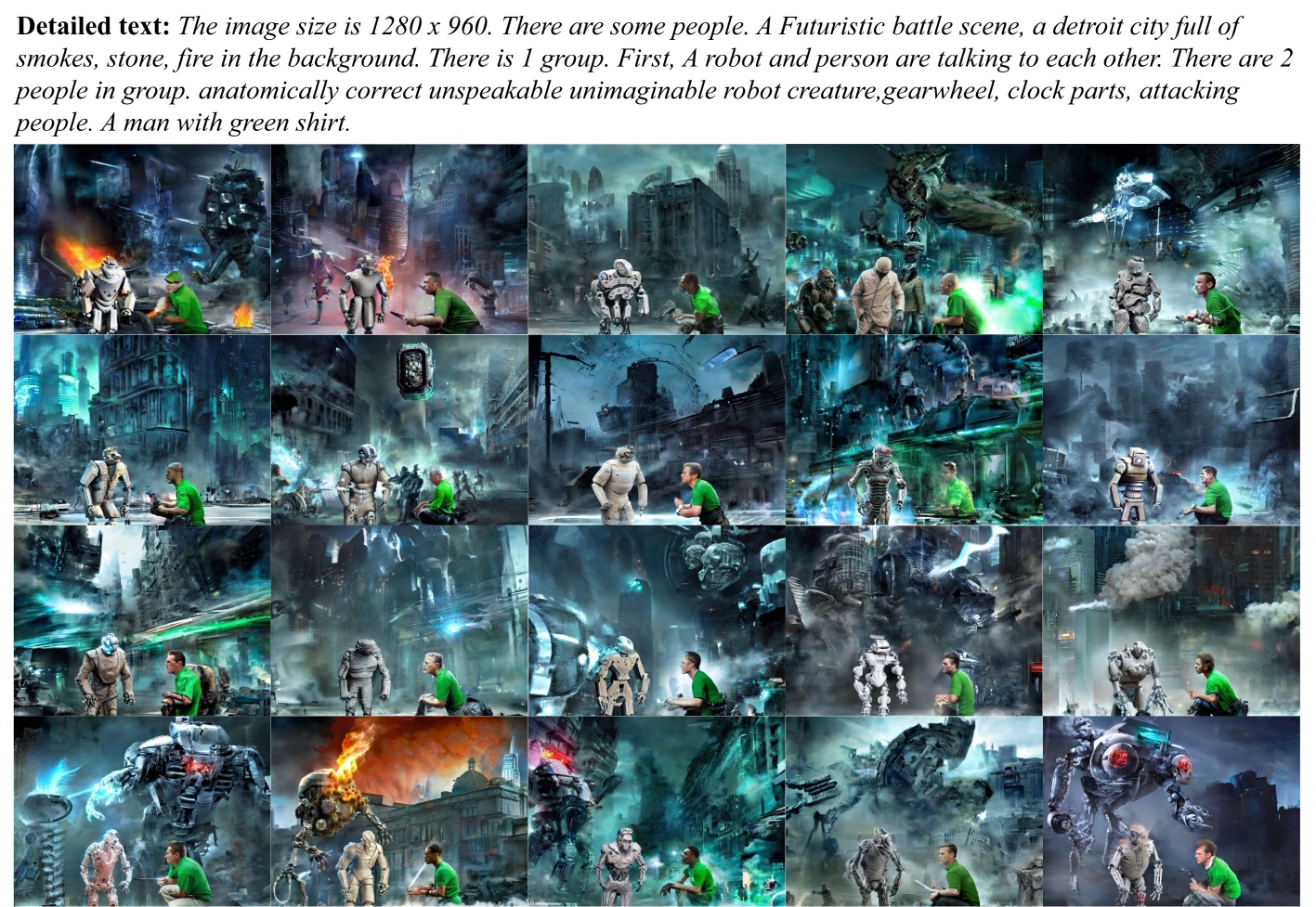}
    \caption{Qualitative results of our DetText2Scene with each detailed text and different seeds(1280$\times$960). Our DetText2Scene generates the diverse large scenes with different seeds while maintaining the high \textit{faithfulness}, \textit{controllability} and \textit{naturalness}.}
    \label{fig_supp_qual4}
\end{figure*}

\begin{figure*}[t]
    \centering
    \includegraphics[width=0.9\textwidth]{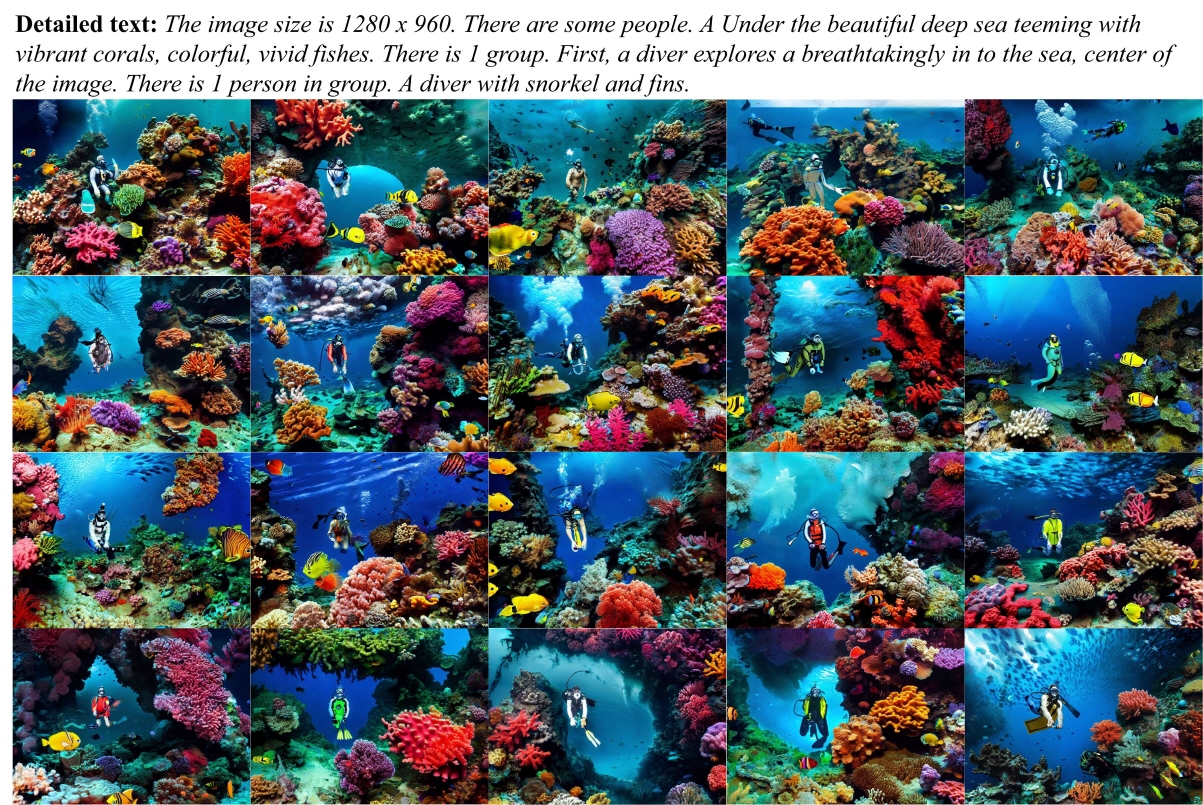}
    \includegraphics[width=0.9\textwidth]{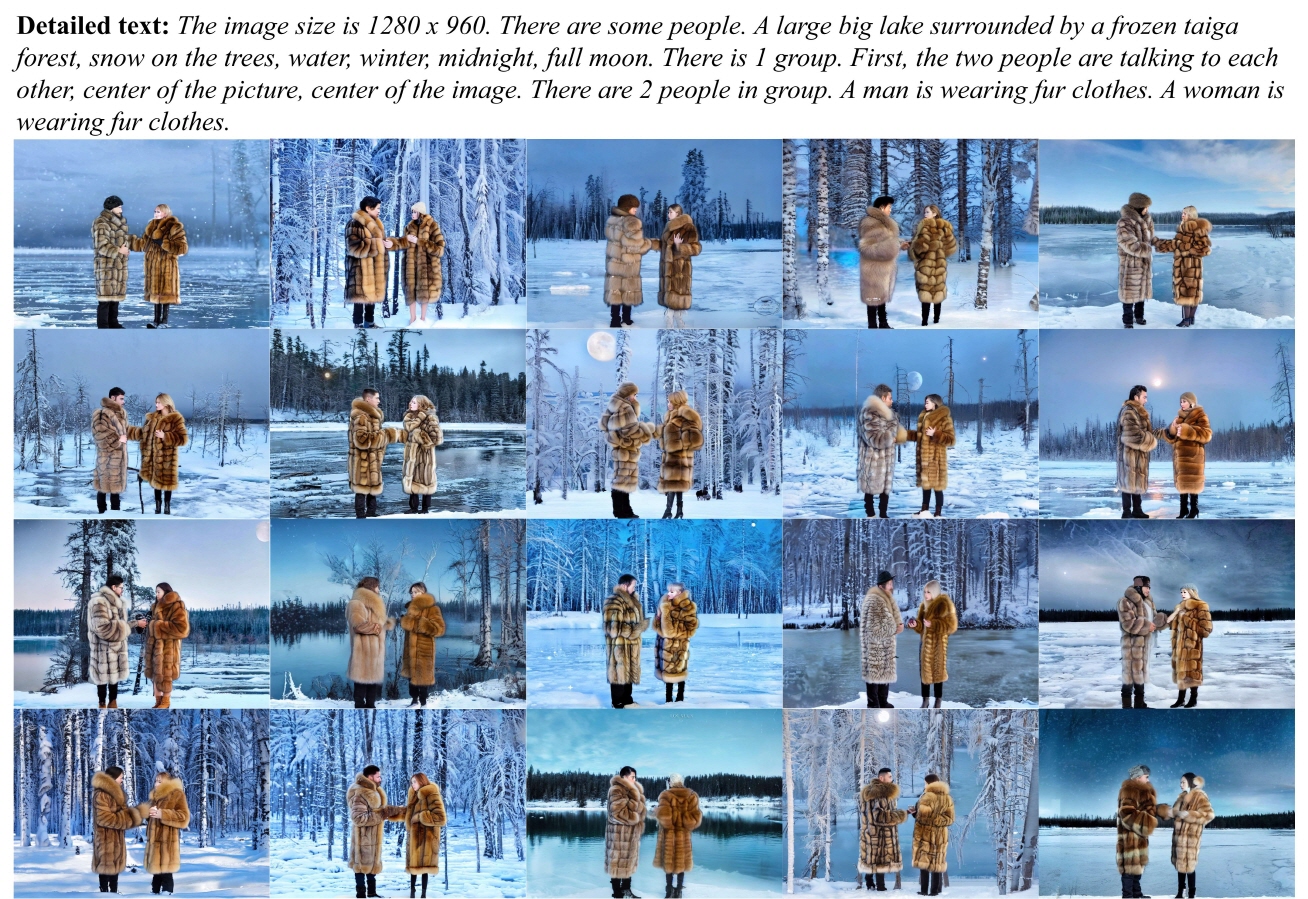}
    \caption{Qualitative results of our DetText2Scene with each detailed text and different seeds(1280$\times$960). Our DetText2Scene generates the diverse large scenes with different seeds while maintaining the high \textit{faithfulness}, \textit{controllability} and \textit{naturalness}.}
    \label{fig_supp_qual5}
\end{figure*}

\begin{figure*}[t]
    \centering
    \includegraphics[width=0.9\textwidth]{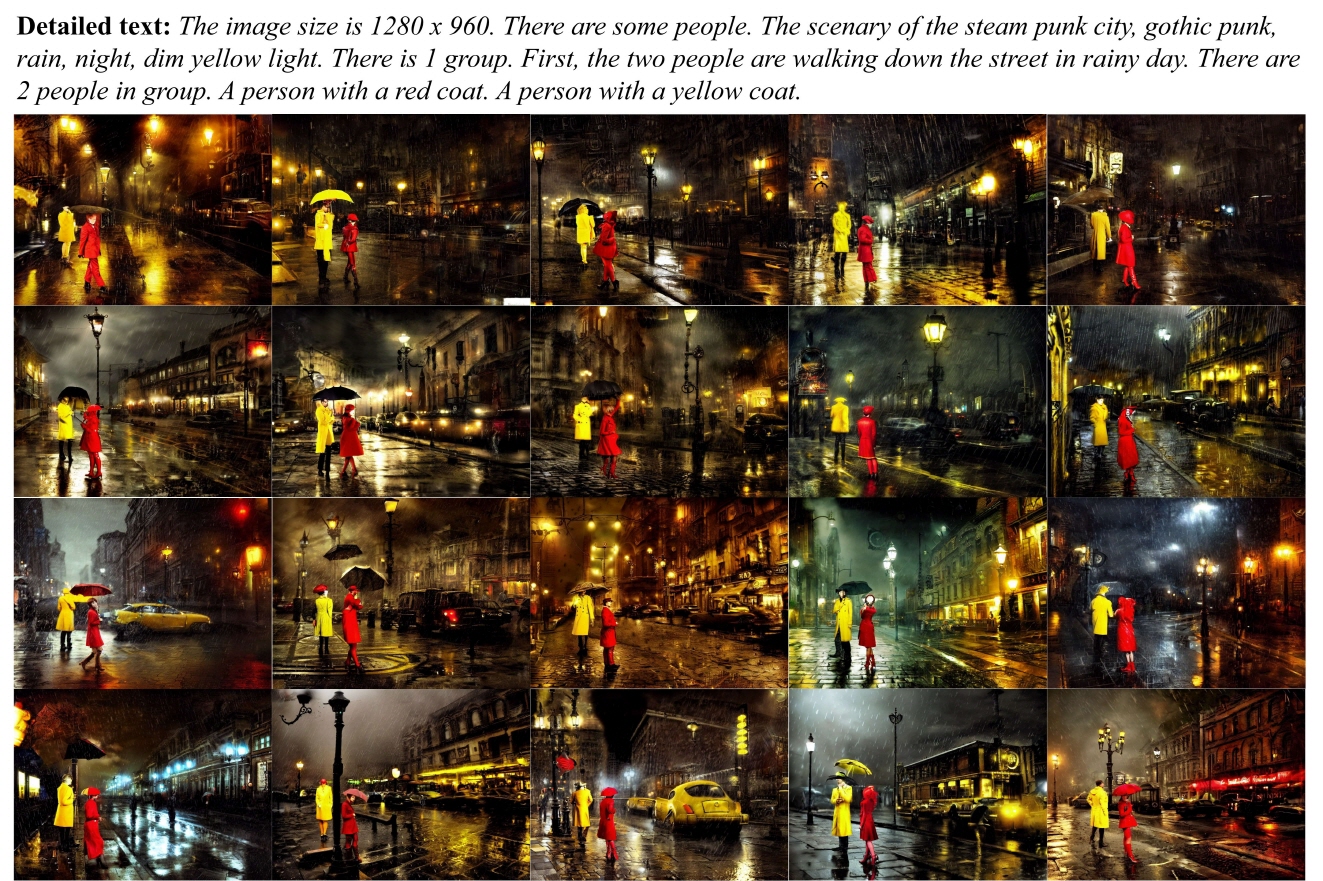}
    \includegraphics[width=0.9\textwidth]{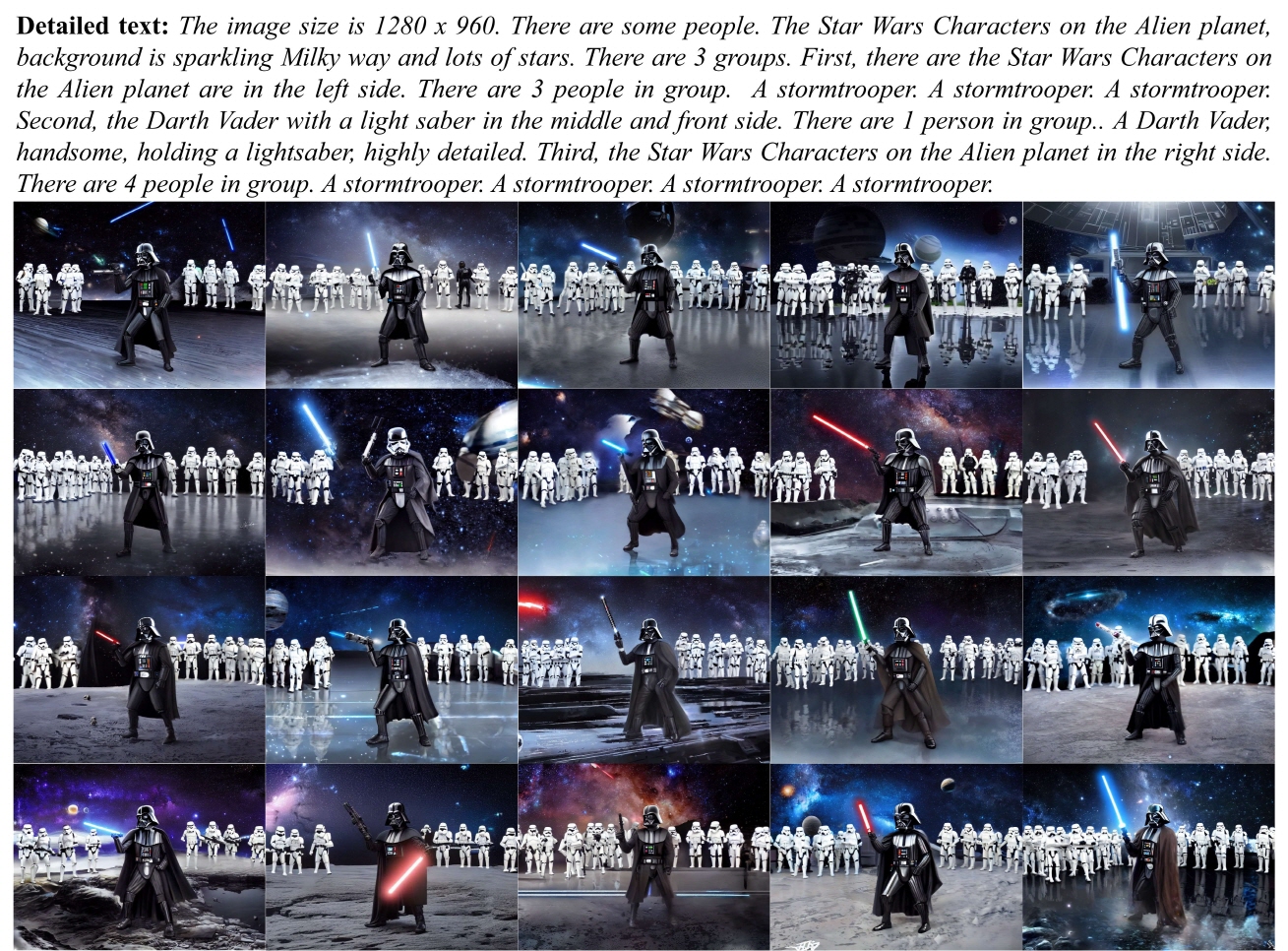}
    \caption{Qualitative results of our DetText2Scene with each detailed text and different seeds(1280$\times$960). Our DetText2Scene generates the diverse large scenes with different seeds while maintaining the high \textit{faithfulness}, \textit{controllability} and \textit{naturalness}.}
    \label{fig_supp_qual6}
\end{figure*}

\begin{figure*}[t]
    \centering
    \includegraphics[width=0.9\textwidth]{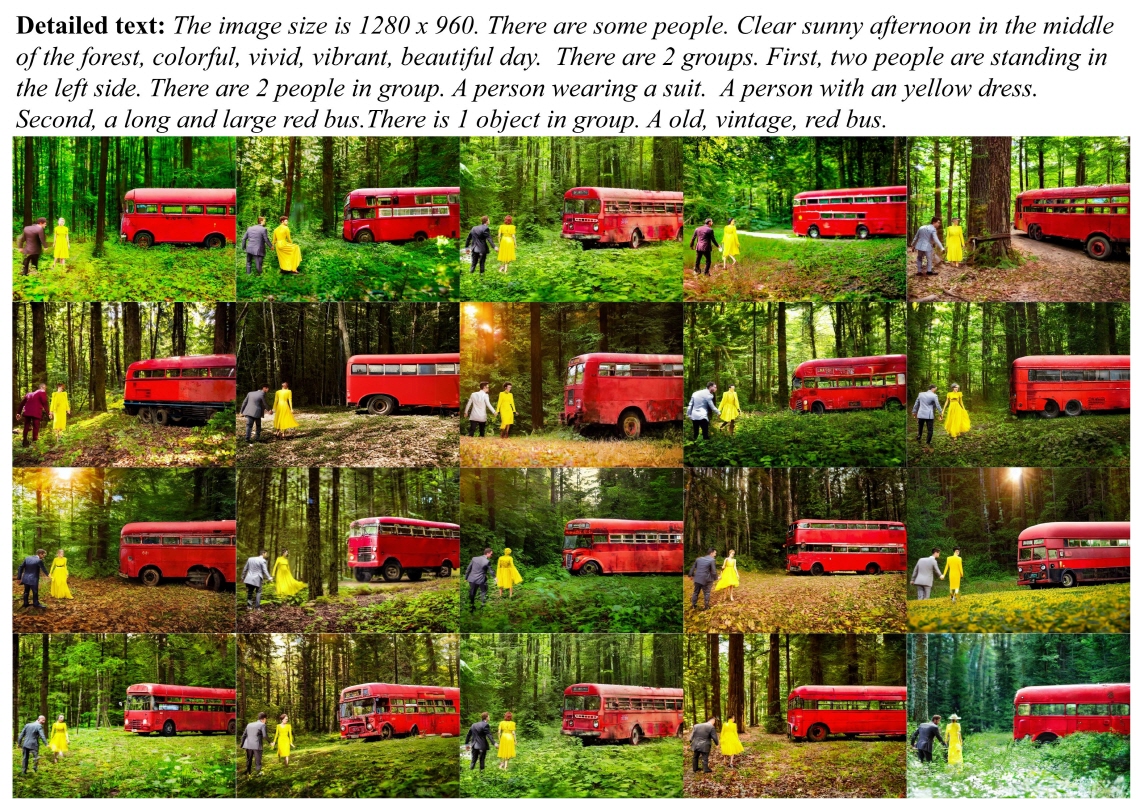}
    \includegraphics[width=0.9\textwidth]{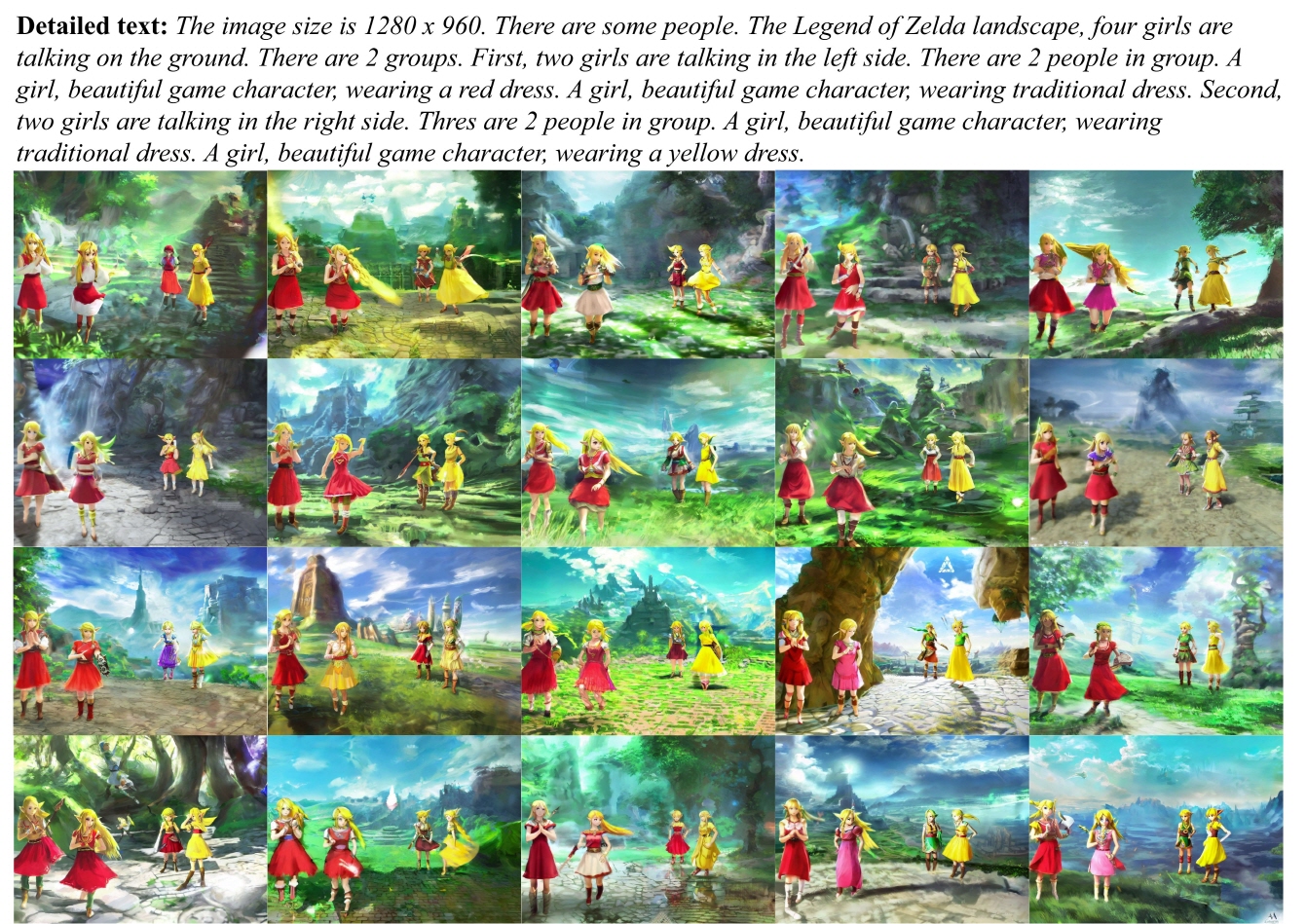}
    \caption{Qualitative results of our DetText2Scene with each detailed text and different seeds(1280$\times$960). Our DetText2Scene generates the diverse large scenes with different seeds while maintaining the high \textit{faithfulness}, \textit{controllability} and \textit{naturalness}.}
    \label{fig_supp_qual7}
\end{figure*}

\begin{figure*}[t]
    \centering
    \includegraphics[width=0.96\textwidth]{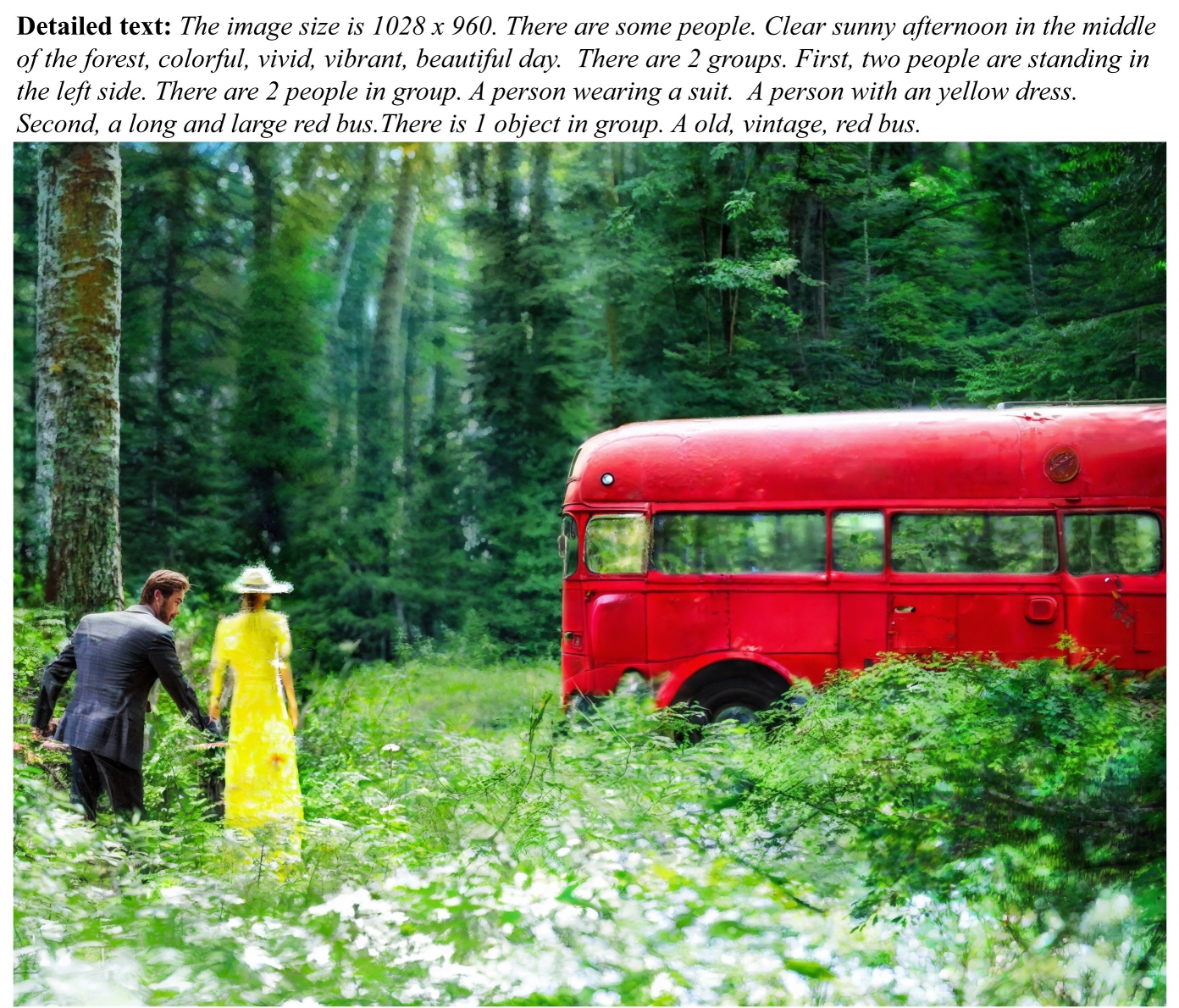}
    \caption{Qualitative main result of our generated large-scaled scene(2560$\times$1920) from text prompt. Through enlarged visualization, it demonstrates that our DetText2Scene properly reflects the \textit{faithfulness}, \textit{controllability} and \textit{naturalness} of detailed text.}
    \label{fig_supp_qual8}
\end{figure*}

\begin{figure*}[t]
    \centering
    \includegraphics[width=0.96\textwidth]{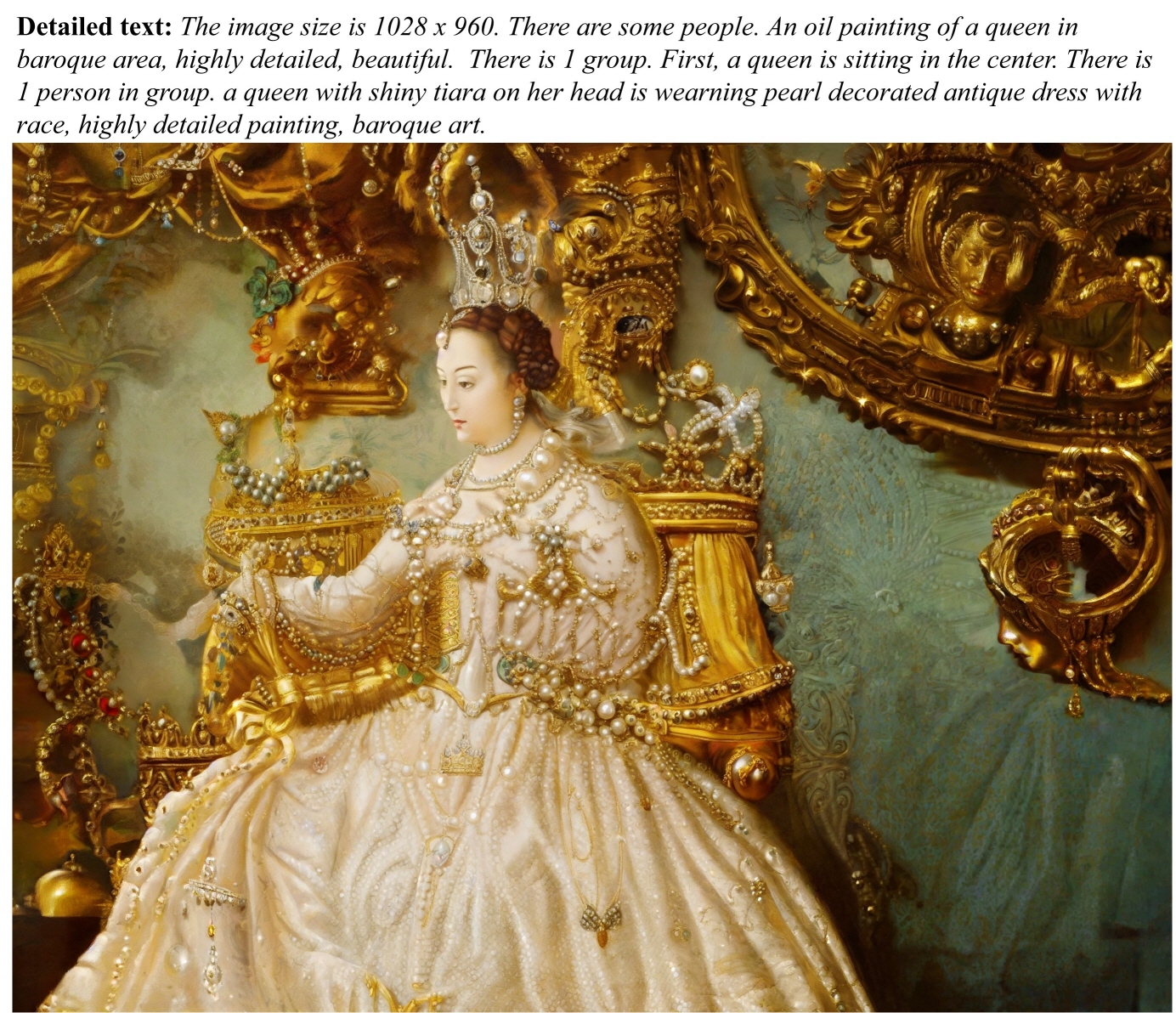}
    \caption{Qualitative main result of our generated large-scaled scene(1028$\times$960) from text prompt. Through enlarged visualization, it demonstrates that our DetText2Scene properly reflects the \textit{faithfulness}, \textit{controllability} and \textit{naturalness} of detailed text.}
    \label{fig_supp_qual9}
\end{figure*}

\begin{figure*}[t]
    \centering
    \includegraphics[width=0.96\textwidth]{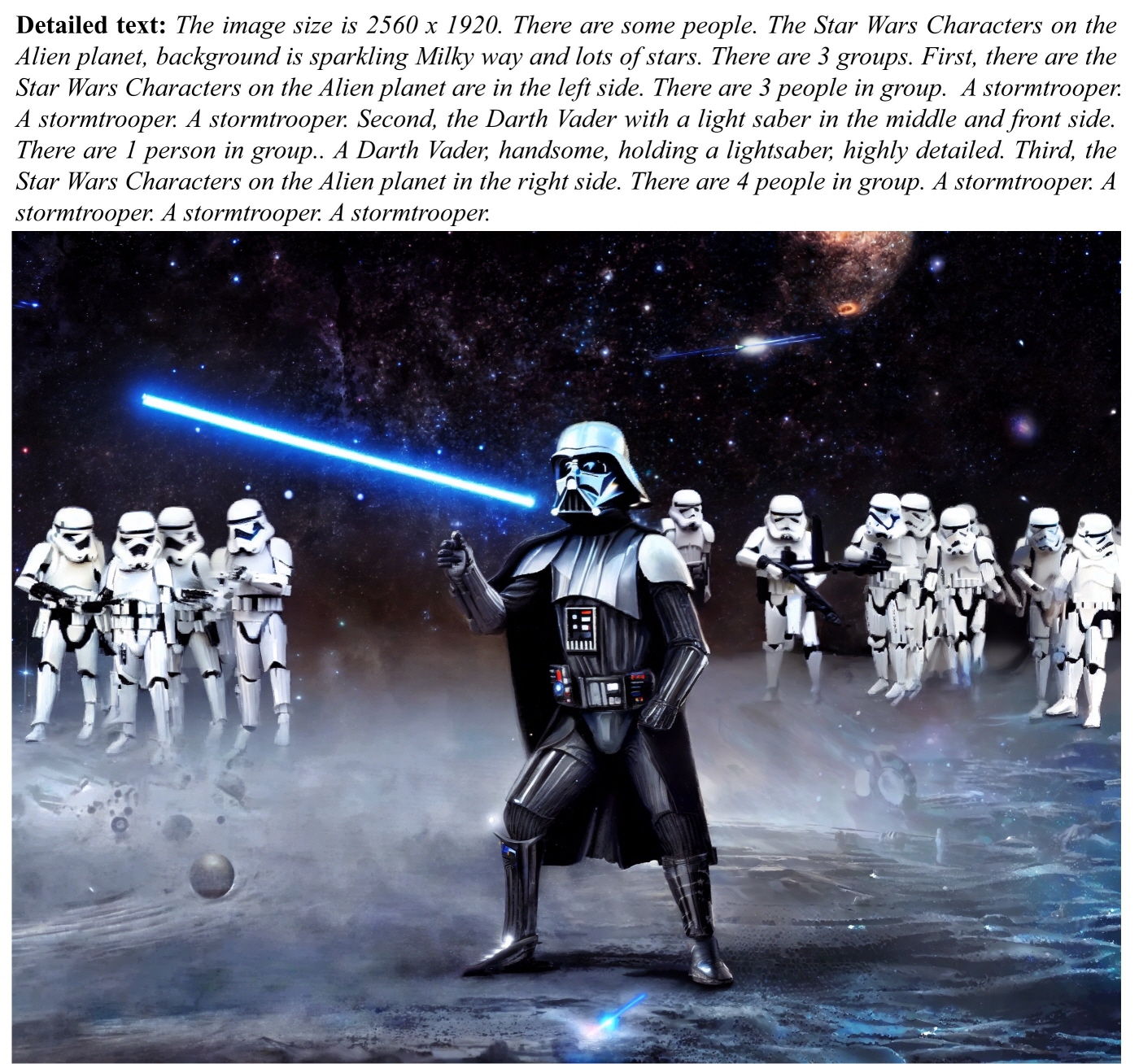}
    \caption{Qualitative main result of our generated large-scaled scene(2560$\times$1920) from text prompt. Through enlarged visualization, it demonstrates that our DetText2Scene properly reflects the \textit{faithfulness}, \textit{controllability} and \textit{naturalness} of detailed text.}
    \label{fig_supp_qual11}
\end{figure*}

\begin{figure*}[t]
    \centering
    \includegraphics[width=0.96\textwidth]{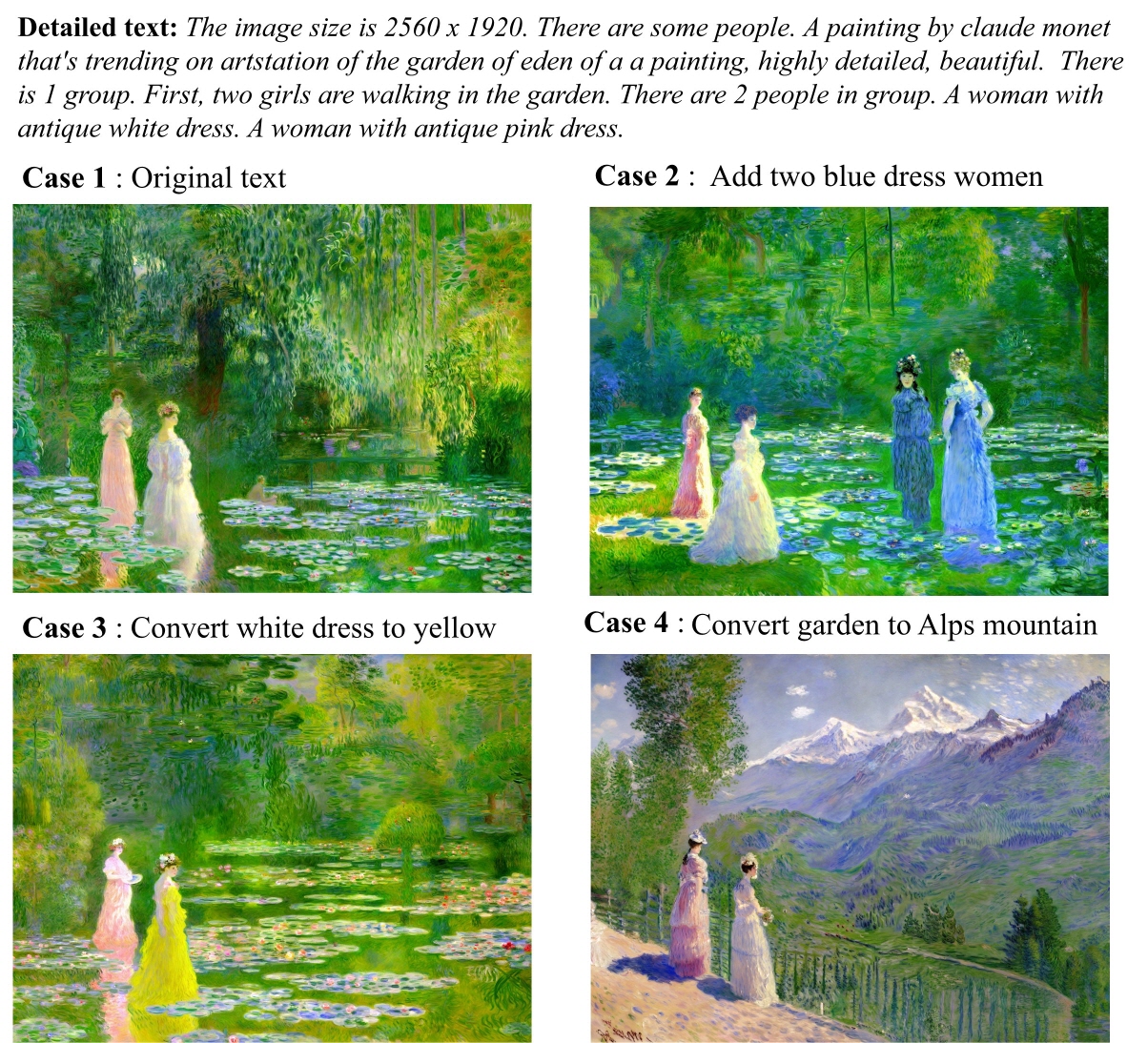}
    \caption{Qualitative results of our DetText2Scene with original and slightly changed detailed text.(2560$\times$1920) We modified three components of detailed text: the background, the number of instances, and the attributes of specific person. Through our DetText2Scene, it demonstrates the synthesis of large-scale scenes controlled by detailed text.}
    \label{fig_supp_qual12}
\end{figure*}

\begin{figure*}[t]
    \centering
    \includegraphics[width=0.96\textwidth]{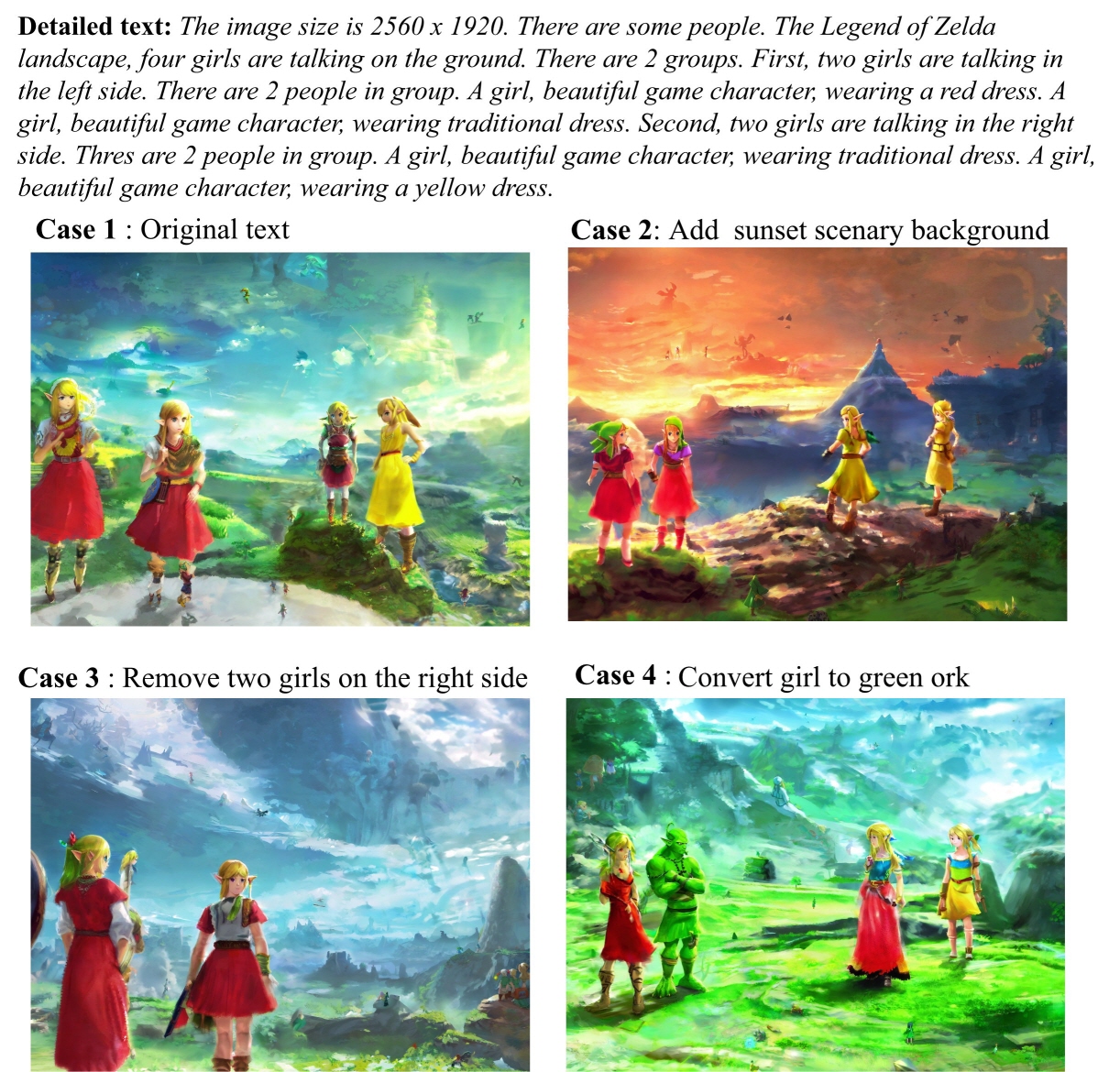}
    \caption{Qualitative results of our DetText2Scene with original and slightly changed detailed text.(2560$\times$1920) We modified three components of detailed text: the background, the number of instances, and the attributes of specific person. Through our DetText2Scene, it demonstrates the synthesis of large-scale scenes controlled by detailed text.}
    \label{fig_supp_qual13}
\end{figure*}

\begin{figure*}[t]
    \centering
    \includegraphics[width=0.96\textwidth]{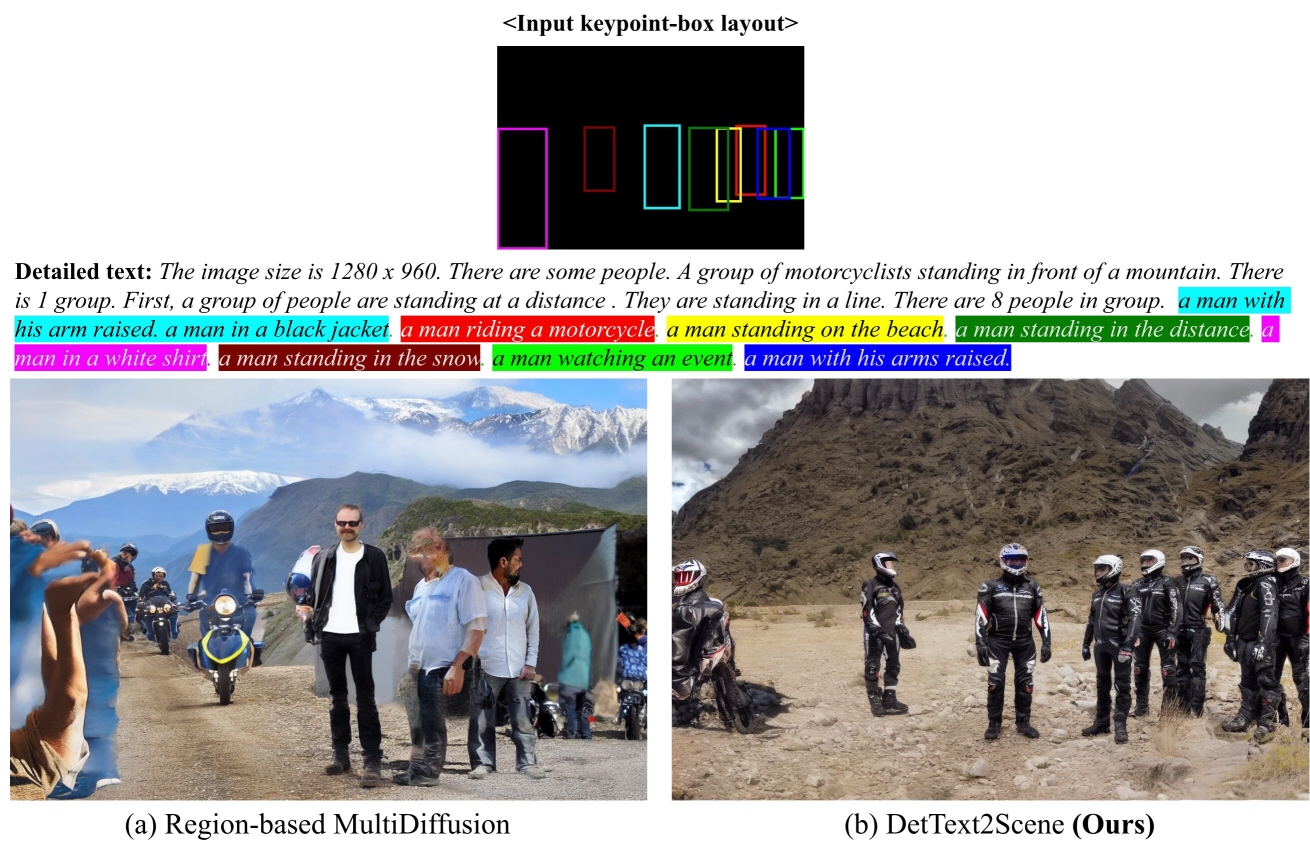}
    \caption{Qualitative results of large-scene image generation from the spatial controls with detailed text.(1280$\times$960) In the Region-based MultiDiffusion~\cite{bar2023multidiffusion}, bounding box masks with instance descriptions were employed to ground each instance. For better visualization, we matched the bounding box with corresponding instance description using distinct colors. It shows that our DetText2Scene successfully generates faithful results that exhibit naturalness in the global context.}
    \label{fig_supp9}
\end{figure*}

\begin{figure*}[t]
    \centering
    \includegraphics[width=0.96\textwidth]{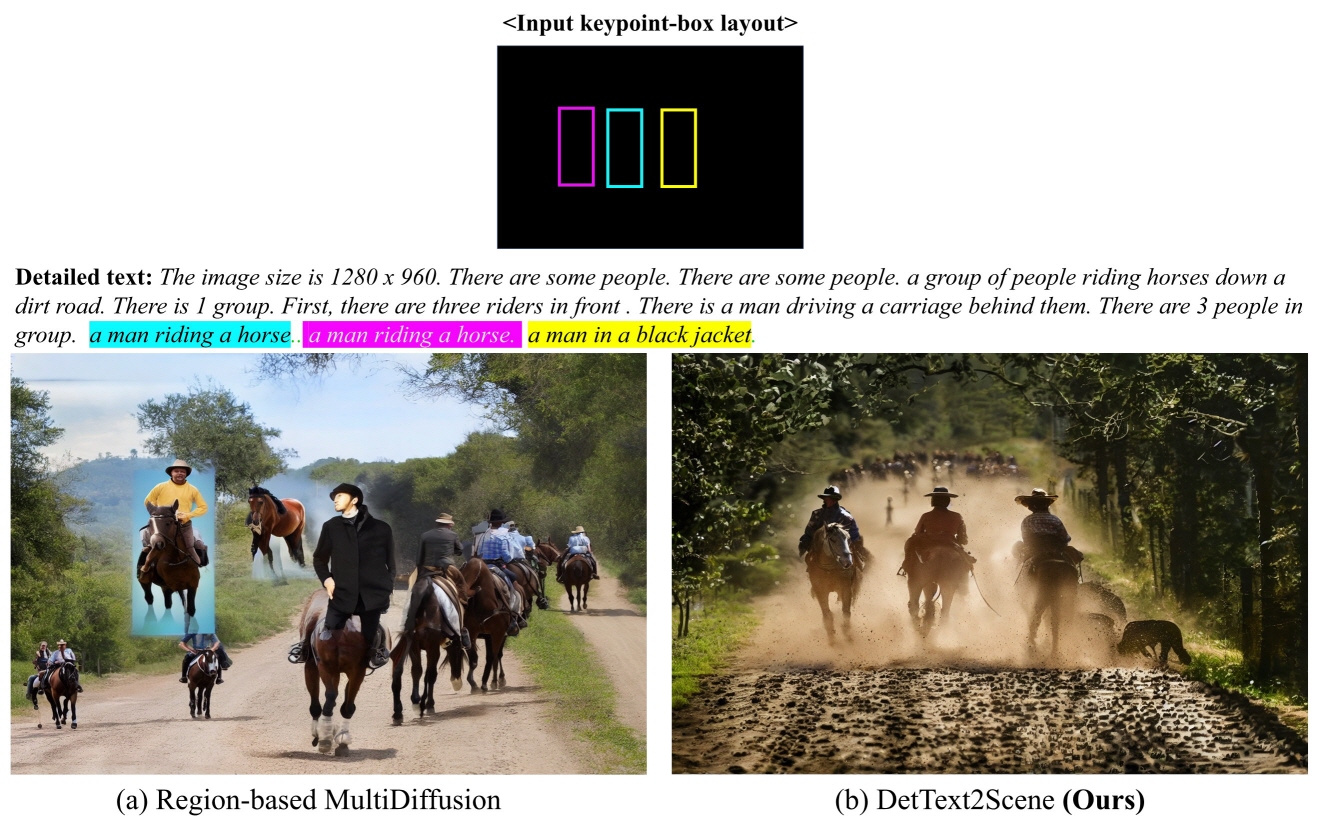}
    \caption{Qualitative results of large-scene image generation from the spatial controls with detailed text.(1280$\times$960) In the Region-based MultiDiffusion~\cite{bar2023multidiffusion}, bounding box masks with instance descriptions were employed to ground each instance. For better visualization, we matched the bounding box with corresponding instance description using distinct colors. It shows that our DetText2Scene successfully generates faithful results that exhibit naturalness in the global context.}
    \label{fig_supp10}
\end{figure*}

\paragraph{User study.} For ablation study of our DetText2Scene, we conducted user study to assess the faithfulness and naturalness of images generated by DetText2Scene w/o PPPI and DetText2Scene w/ PPPI. The details of user study are same as Section~\ref{B3_largeimagegen}. As shown in the Table~\ref{tab3_ablation}, the DetText2Scene w/ PPPI were significantly preferred in terms of faithfulness and naturalness compared to the the DetText2Scene w/o PPPI.
Here, we present some examples of the questionnaires of user study comparing our DetText2Scene with PPPI and without PPPI. in Fig.~\ref{fig_supp_user_pppi}.

\end{document}